\documentclass[smallextended]{svjour3}

\usepackage{hyperref}
\usepackage[usenames,dvipsnames]{xcolor}
\usepackage[framemethod=TikZ]{mdframed}
\usepackage{multicol}
\usepackage{changes}
\usepackage{caption}
\usepackage{graphicx}
\usepackage{url}
\usepackage{todonotes}
\usepackage{amsmath}
\usepackage{amssymb}
\usepackage{bm}
\usepackage{multirow}
\usepackage{paralist}
\usepackage{amsfonts}
\usepackage{enumitem}
\usepackage{tcolorbox}
\usepackage{booktabs}
\usepackage{tabularx}
\usepackage{rotating}
\usepackage{marginnote}
\usepackage{subfig}
\usepackage{makecell}

\begin{document}




\title{Fire Now, Fire Later: Alarm-Based Systems for Prescriptive Process Monitoring}

%

\author{Stephan A. Fahrenkrog-Petersen \and Niek Tax \and Irene Teinemaa \and Marlon Dumas \and Massimiliano de Leoni \and Fabrizio Maria Magg \and Matthias Weidlich}

\institute{S. Fahrenkrog-Petersen \and M. Weidlich \at
              Humboldt-Universität zu Berlin, Berlin, Germany \\
              \email{\{stephan.fahrenkrog-petersen, matthias.weidlich\}@hu-berlin.de} \\
            \and I. Teinemaa \and M. Dumas \at
          	University of Tartu, Tartu, Estonia\\
          	\email{\{irene.teinemaa, marlon.dumas\}@ut.ee} \\
          	\and N. Tax \at
          	TU Eindhoven, Eindhoven, The Netherlands\\
          	\email{niektax@gmail.com} \\
          	 \and M. de Leoni  \at
          	University of Padua, Padua, Italy\\
          	\email{deleoni@math.unipd.it} \\
          	\and F. Maggi  \at
          	Free University of Bozen-Bolzano, Bolzano, Italy\\
          	\email{maggi@inf.unibz.it}          
      }

\maketitle

\begin{abstract}
Predictive process monitoring is a family of techniques to analyze events
produced during the execution of a business process in order to predict the
future state or the final outcome of running process instances.
Existing techniques in this field are able to predict, at each step of a
process instance, the likelihood that it will lead to an undesired outcome.
These techniques, however, focus on generating predictions and do not prescribe
when and how process workers should intervene to decrease the cost of undesired
outcomes.
This paper proposes a framework for prescriptive process monitoring, which
extends predictive monitoring with the ability to generate alarms that
trigger interventions to prevent an undesired outcome or mitigate its
effect.
The framework incorporates a parameterized cost model to assess the
cost-benefit trade-off of generating alarms.
We show how to optimize the generation of alarms given an event
log of past process executions and a set of cost model parameters.
The proposed approaches are empirically evaluated using a range of real-life
event logs.
The experimental results show that the net cost of undesired outcomes can be
minimized by changing the threshold for generating alarms, as the process
instance progresses. Moreover, introducing delays for triggering alarms, instead of triggering them as soon as the probability of an undesired outcome exceeds a threshold, leads to lower net costs.
\end{abstract}




\section{Introduction}
\label{sec:intro}
The general idea of process mining~\cite{DBLP:books/sp/Aalst16} is to discover, monitor and improve real-life business processes by extracting knowledge from event logs of past process executions \cite{AugustoCDRP19,LeemansF20,ZelstDA18} recorded by different systems that gather event data. Over the last ten years, event data have become more widely available and process mining techniques have greatly matured.

\emph{Predictive process
monitoring}~\cite{maggi2014predictive,MetzgerLISFCDP15} is a family of process mining
techniques to predict the future state of running instances of a business
process based on event logs.
Respective techniques may provide predictions on the remaining execution time
of a process instance (herein called a \emph{case}), the next activity to be
executed, or the final outcome of the process instance with respect to a set of
possible outcomes. This article is concerned with the latter type of predictive
process monitoring, which we call
\emph{outcome-oriented}~\cite{teinemaa2017outcome}. For example, in a
lead-to-order process, outcome-oriented techniques may predict
whether a case will lead to an order (desired outcome), or not (undesired
outcome).

Existing techniques for outcome-oriented predictive process monitoring are able
to predict, after each event of a case, the probability that the case will end
with an undesired outcome. Yet, these techniques are restricted
to prediction, not suggesting or prescribing how and when
process workers should intervene to decrease the probability or cost of
undesired outcomes.

A naive approach to turn a predictive process monitoring technique into a
prescriptive one is by triggering an alarm when the probability that a case will
lead to an undesired outcome is above a fixed threshold (e.g.,\ 90\%). This
alarm can be linked to an intervention, such as calling the customer, offering
a discount, etc. Yet, this naive approach may be far from optimal, as
interventions induce a cost (e.g.,\ time spent by workers in the
intervention or forgone revenue) and an effect (e.g.,\ preventing the
undesired outcome altogether or only partially mitigating it). Moreover, the
cost and the effect of an intervention may vary as the case advances.
Late interventions may be less effective and more costly than
earlier interventions, in such a way that the net cost of intervening varies
(possibly non-monotonically) as the case advances.

This article proposes a framework to extend predictive process monitoring techniques in order to make them prescriptive.
The proposed framework extends a given predictive process monitoring model with a mechanism for generating alarms that lead to interventions, which, in turn, mitigate (or altogether prevent) undesired outcomes.
The framework is armed with a cost model that captures,
among others, the trade-off between the cost of an intervention and the cost of
an undesired outcome.
Based on this cost model, the article presents an approach to tune the generation of alarms so as to minimize the expected cost for a given event log and a set of parameters.
The approach is empirically evaluated, under various configurations, using a collection of real-life event logs.

This article is an extended and revised version of a previous conference
paper~\cite{Teinemaa2018Prescriptive}. The conference version focused on the
scenario where there is a single type of alarm leading to a single type of
intervention (e.g., calling the customer). This article enhances the scope of
the framework to consider multiple alarm types, each one leading to a different
intervention. For example, one alarm type may lead to calling the customer while another
one leads to offering a
discount by email. Additionally, this article considers further factors that
influence the effectiveness of alarms in practice. First, we
add the possibility that the probability threshold above which an alarm is
triggered may vary depending on how far the case has progressed. Second, we
introduce the possibility of delaying the firing of an alarm to reduce
false alarms stemming from instability in the predictive model.

The article is structured as follows.
\autoref{sec:related} discusses related work.
Next, \autoref{sec:framework} presents the
prescriptive process monitoring framework. \autoref{sec:approach} outlines
the approach to optimize the alarm generation mechanism, while
\autoref{sec:evaluation} reports on the empirical evaluation. \autoref{sec:conclusion} concludes the paper and spells out directions for future work.

\section{Related Work}
\label{sec:related}
As stated above, a naive approach to turn a model for predicting undesired process execution outcomes into a prescriptive model is to raise an alarm whenever the predictive model estimates that the probability of a negative outcome is above a given threshold. This, in turn, raises the question of determining an optimal alarm threshold.
The problem of determining an optimal alarm threshold with respect to a given
cost function is closely related to the problem of \emph{cost-sensitive
learning}.
Cost-sensitive learning seeks to find an optimal prediction when different
types of misclassifications have different costs
and different types of correct
classifications have different benefits~\cite{elkan2001foundations}.
A non-cost-sensitive classifier can be turned into a cost-sensitive one by
\emph{stratification} (rebalancing the ratio of positive and
negative training samples)~\cite{elkan2001foundations}, by learning a
\emph{meta-classifier} after relabeling the training samples according to their
estimated cost-minimizing class label~\cite{domingos1999metacost}, or via
empirical thresholding~\cite{sheng2006thresholding}. In this article, we adopt the latter approach, which has been shown to perform better than other
cost-sensitive learning approaches~\cite{sheng2006thresholding}.

While the above-cited cost-sensitive learning approaches provide a starting point for turning prediction models (e.g., classifiers) for triggering alarms, they are not designed to tackle the problem of prescriptive process monitoring. In particular, the above approaches target the scenario where the predictions made by a model are immediately used to make a decision (e.g., triggering an intervention). In particular, these approaches do not consider the possibility of
delaying the decision. In this article, we deal with a different problem formulation,
where the costs may depend on the time when the decision is
made, i.e., one may delay the decision to accumulate further
information.

Cost-sensitive learning for sequential decision-making has been
approached using reinforcement learning (RL)
techniques~\cite{pednault2002sequential}. This work differs from ours in three
ways. First, instead of making a sequence of decisions, we aim at finding an
optimal time to make a single decision (the decision to raise an alarm).
Second, RL assumes that actions affect the observed state once they are triggered. In our case, there are two possible actions at each step: 1)
raising one of the available alarms, or 2) delaying the decision. In the latter case, we will wait until we observe
the next event. But this event is not
at all affected by the selected action (i.e., to ``wait'').
In other words, the environment is not affected by the delay in the decision.
Finally, we train our model only based on observed data, while reinforcement learning requires the existence of a simulator, that imitates the environment, in our case the business process.

Predictive and prescriptive process monitoring are also related to Early
Classification of Time Series (ECTS), which aims at accurate classification of
a (partial) time series as early as possible~\cite{xing2012early}.
Common solutions for ECTS find an optimal
\emph{trigger function} that decides on whether to output the prediction or to
delay the decision and wait for another observation in the time series. To this
end, the approach presented in \cite{xing2012early} identifies the minimum prediction length when the
memberships assigned by the nearest neighbor classifier become stable, the one presented in \cite{parrish2013classifying} estimates the probability that the label assigned based on the current prefix is the same as the one assigned based on the
complete time series, and, similarly, the one presented in \cite{mori2017early,mori2017reliable} compares the accuracy achieved based on a prefix to the one achieved based on a complete trace.
Recently, a few non-myopic methods have been
proposed~\cite{dachraoui2015early,tavenard2016cost}. Yet, these approaches
assume a-priori knowledge of the length of the sequence, which is not given in
the context of traces of business processes. As such, the problem is different,
as delaying the decision comes with
a risk that the case will end before the next possible decision point.
The methods outlined
in~\cite{mori2017early,mori2017reliable,dachraoui2015early,tavenard2016cost} are the
only ECTS methods that try to balance
accuracy-related and earliness-related costs.
However, they assume that predicting a positive class early has the same effect on the
cost function as predicting a negative class early, which is not the case in
typical business process monitoring scenarios, where earliness matters only
when an undesired outcome is predicted.

We are aware of five previous studies related to alarm-based prescriptive
process
monitoring~\cite{metzger2017predictive,di2016clustering,DBLP:conf/caise/MetzgerNBP19,groger2014prescriptive,krumeich2016prescriptive}.
\cite{metzger2017predictive} study the effect of different
likelihood thresholds on the total intervention cost (called adaptation
cost) and the misclassification penalties. However, they assume that the alarms
can only be generated at one pre-determined point in the process that can be
located as a state in a process model. Hence, their approach is restricted to
scenarios where: (i) there is a process model that perfectly captures all
cases; (ii) the costs and rewards implied by alarms are not time-varying.
Also, their approach relies on a mechanism with a user-defined threshold, as
opposed to our empirical thresholding approach. The latter remark also applies
to~\cite{di2016clustering,DBLP:conf/caise/MetzgerNBP19}, which generate a prediction when the
likelihood returned by a trace prefix classifier first exceeds a given
threshold. Also, the study proposed in~\cite{di2016clustering} is not cost-sensitive, whereas the one introduced in~\cite{DBLP:conf/caise/MetzgerNBP19} is based on a static cost model that does not change over time.
\cite{groger2014prescriptive} provide recommendations during
the execution of business processes to avoid a predicted performance deviation.
Yet, this approach
does not take into account the notion of earliness, i.e., the fact that
firing an alarm earlier has a different effect than firing it at a later point.
Finally, \cite{krumeich2016prescriptive} propose a general
architecture for prescriptive process monitoring. However, this architecture
does not incorporate an alarm model nor does it propose a method for cost
optimization.

\section{Prescriptive Process Monitoring Framework}
\label{sec:framework}

This section introduces a framework for an alarm-based prescriptive process
monitoring system.
We first introduce the notion of event log in Section~\ref{sec:background},
before turning to the cost model of our framework in
Section~\ref{sec:singlecosts}. For the sake of clarity, this model is first
introduced for scenarios that allow for one type of alarm. The latter
assumption is dropped in
Section~\ref{sec:multicosts} extending the cost model to scenarios with
multiple alarms. Finally, Section~\ref{sec:costInstances} builds on the
cost-model to formalize the concept of an alarm system.

\subsection{Event Log}
\label{sec:background}

\begin{figure}[t]
\centering
  \includegraphics[width=.85\textwidth]{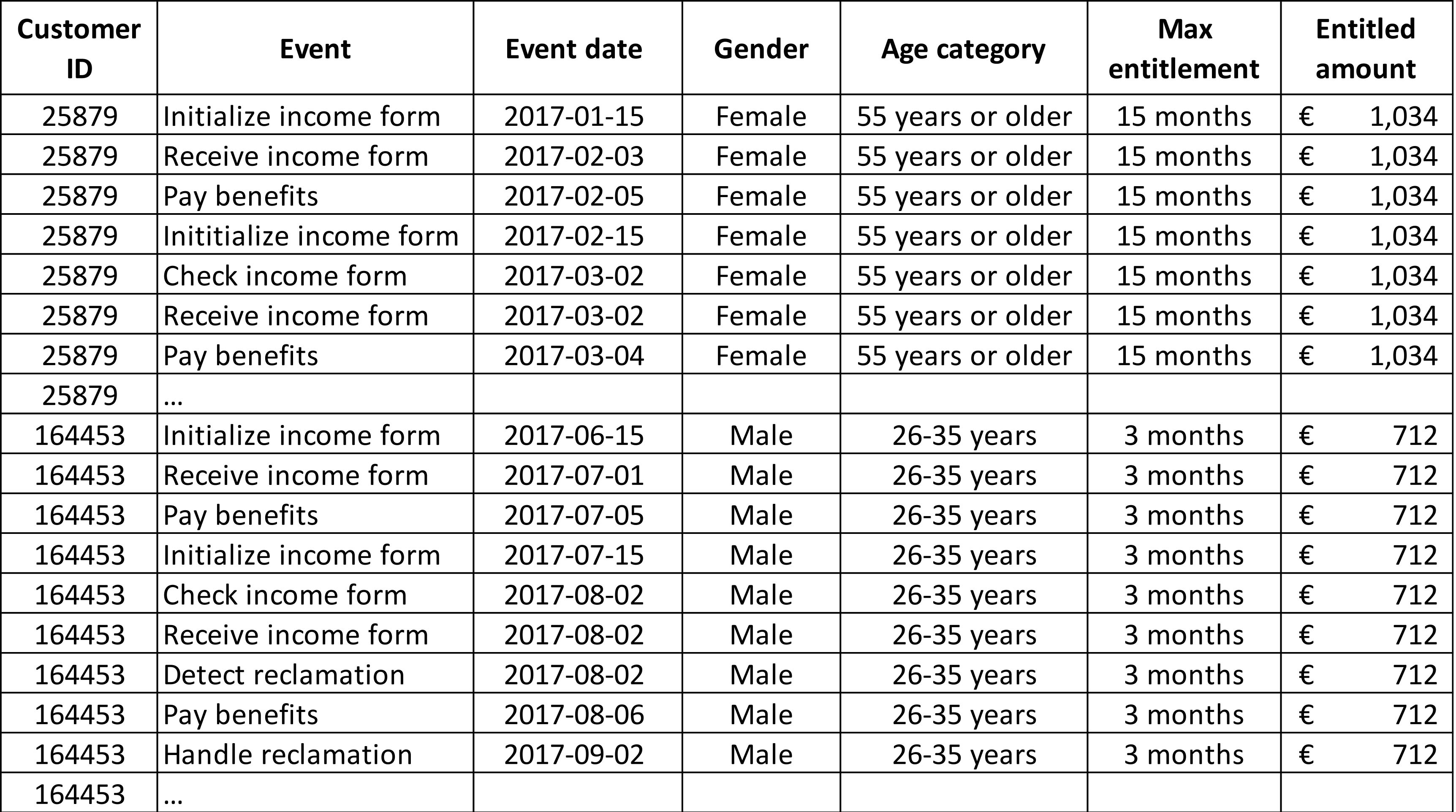}
  \caption{A fragment of an event log for two UWV's customers, as
  explained in Example~\ref{ex:unemployment}. Each row is an event;
  events with the same customer id are grouped into traces.
}
  \label{fig:eventLog}
\end{figure}

Executions of process instances (a.k.a. cases) are recorded in so-called \emph{event logs}.
At the most abstract level, an event log can be represented in a tabular form.
Figure~\ref{fig:eventLog} shows an excerpt of an event log that refers to the execution of the unemployment-benefit process in the Netherlands.
Example~\ref{ex:unemployment} describes how the process is usually executed,
which serves for illustration in the remainder of the paper.
The description is the result of several interactions of one of the authors (see, e.g.,~\cite{D_dL_M@COOP17}) with the company's stakeholders.
\begin{example}[Unemployment Benefit]
\label{ex:unemployment}
In the Netherlands, UWV is the institution that provides
social-security insurances for Dutch residents.
UWV provides several insurances. One of the most relevant insurances is the provision of unemployment benefit.
When residents (hereafter customers) become unemployed, they are usually
entitled to monthly monetary benefits for a certain period of time.
UWV executes a process to determine the amount of these benefits, to manage the interaction with the customers, and to perform the monthly payments.
These payments are stopped when the customer reports that he/she has found a new job or when the time period in which the customer is entitled to unemployment benefits ends.
\end{example}
\noindent Each row in Figure~\ref{fig:eventLog} corresponds to an event. An
event represents the execution of an activity for a case with a certain
identifier that occurred at a specific moment in time and was performed by a
given resource. The case identifier can vary depending on the process: for the
example in question, it coincides with the customer id.
Events can be grouped by case identifier and ordered by timestamp, thus obtaining a sequence of events, a.k.a.\ a trace.

Figure~\ref{fig:eventLog} shows that events can be associated with various
properties, namely the different columns of the event table; we assume
that the activity name and the timestamp are properties that are always
present. Given an event $e$ of a log, $\pi_\mathcal{T}(e)$ and
$\pi_\mathcal{A}(e)$ return the timestamp and the activity name
of an event $e$, respectively.
As an example, the trace of customer id 25879 in Figure~\ref{fig:eventLog} is
$\langle e_1,e_2,\ldots,e_n \rangle$ where, e.g., it holds that
$\pi_\mathcal{T}(e_1)=\texttt{2017-01-15}$ and
$\pi_\mathcal{A}(e_1)=\textit{`Initialize income form'}$.
In the remainder, $\langle\rangle$ is the empty sequence, and $\sigma_1 \cdot \sigma_2$ indicates the concatenation of sequence $\sigma_1$ and $\sigma_2$. Given a sequence $\sigma=\langle a_1,a_2,\dots,a_n\rangle$, for any $0<k\leq n$, $\mathit{hd}^k(\sigma)=\langle a_1, a_2, \dots, a_k\rangle$ is the prefix of length $k$ ($0 < k < n$) of sequences $\sigma$, and $\sigma(k)=a_k$ is the $k$-th event of $\sigma$.

With these concepts at hand, a trace is defined as a sequence of events, and an event log as a set of traces:
\begin{definition}[Trace, Event Log]
Let $\mathcal{E}$ be the universe of all events.
	A \emph{trace} is a finite non-empty sequence of events
	$\sigma\in\mathcal{E}^+$ such that each event appears at most once and time
	is non-decreasing, i.e., for $1\le i < j \le |\sigma|:\sigma(i)\neq\sigma(j)$
	and $\pi_\mathcal{T}(\sigma(i))\le\pi_\mathcal{T}(\sigma(j))$. An \emph{event
	log} is a set of traces $L\subset\mathcal{E}^+$ such that each event appears
	at most once in the entire log.\looseness=-1
\end{definition}

\newcommand{\true}{\textsf{true}}
\newcommand{\false}{\textsf{false}}
\newcommand{\avg}{\textsf{avg}}

\newtheorem{exampleCont}{Example}

\subsection{Single-Alarm Cost Model}
\label{sec:singlecosts}
An alarm-based prescriptive process monitoring system (\emph{alarm system} for
short) is a monitoring system that can raise an alarm in relation to a running
case of a business process in order to indicate that the case is likely to lead
to some undesired outcome. These alarms are handled by process workers who
intervene in the process instance by performing an action (e.g., by calling a
customer or blocking a credit card), thereby preventing the undesired outcome
or mitigating its effect. These actions may have a cost, which we call
\emph{cost of intervention}. When a case does end in a negative outcome, this
leads to the \emph{cost of undesired outcome}. Clearly, intervening
is beneficial if its cost is lower than the cost of reaching an undesired
outcome.

For many practical application scenarios, however, it is not sufficient to
consider solely a trade-off between the cost of intervention and the cost of
undesired outcome. This happens when interventions that turn
out to be superfluous induce further costs, ranging from financial damage to
losing customers.
We capture these effects by a \emph{cost
of compensation} in case the intervention was unnecessary, i.e., if the case
was only suspected to lead to an undesired outcome, but this suspicion was
wrong.

A further aspect to consider is the
\emph{mitigation effectiveness}, i.e., the relative benefit of raising an alarm
at a certain point in time. Consider again the process of handling unemployment
benefits at UWV as detailed in Example~\ref{ex:unemployment}. In case of
unlawful benefits, the longer UWV postpones an intervention, the lower
its effectiveness, since the amount paid already to the customer
cannot always be claimed back successfully.

Having introduced the various costs to consider when interfering in the
execution of a process instance, we need to clarify the
granularity at which these costs are modeled. In general, an alarm system is
intended to continuously monitor the cases of the business process. However,
in many application scenarios, such continuous monitoring is expensive
(resources to interfere need to be continuously available). Hence, an
approximation may be employed that assumes that significant cost changes are
always correlated with the availability of new events (i.e., new information)
in a case. Following this
line,
alarms can only be raised after the occurrence of an event.

In the remainder, each case is identified by a trace $\sigma$ that is
(eventually) recorded in an event log. For a prefix of such a trace, the
above characteristics of an alarm, i.e., its cost of intervention,
cost of compensation, and mitigation effectiveness, are captured by an alarm
model. These characteristics may depend on the position in the
case at which the alarm is raised and/or on other cases being executed. Hence,
they are defined as functions over the number of already recorded events and
the entire set of cases being executed, as follows:

\begin{definition}[Alarm Model]
\label{def:alarmcosts}
An \emph{alarm model} is a tuple $(c_\mathit{in},c_\mathit{com},\mathit{eff})$ consisting of:
\begin{itemize}[noitemsep,topsep=0pt]
\item a function $c_\mathit{in}\in\mathbb{N}\times\mathcal{E}^+\times 2^{\mathcal{E}^+}\rightarrow \mathbb{R}^+_0$ modeling the \emph{cost of \underline{in}tervention}:
given a trace $\sigma$ belonging to an event log $L$, $c_\mathit{in}(k,\sigma,L)$ indicates the cost of an intervention for a running case with trace $\sigma$ when the intervention takes place after the $k$-th event;

\item a function $c_\mathit{com}\in\mathcal{E}^+\times 2^{\mathcal{E}^+}\rightarrow \mathbb{R}^+_0$ modeling the \emph{cost of \underline{com}pensation};

\item a function $\mathit{eff} \in\mathbb{N}\times\mathcal{E}^+\times 2^{\mathcal{E}^+}\rightarrow[0,1]$ modeling the \emph{mitigation \underline{eff}ectiveness} of an intervention:
for a trace $\sigma$ of an event log $L$, $\mathit{eff}(k,\sigma,L)$ indicates
the mitigation effectiveness of an intervention in $\sigma$ when the
intervention takes place after the k-th event.
\end{itemize}
\end{definition}

\noindent
Combining the above properties of an alarm with the cost of the
undesired outcome that shall be prevented, we define a
single-alarm cost model.

\begin{definition}[Single-Alarm Cost Model]
\label{def:singlecosts}
A \emph{single-alarm cost model} is a tuple
$(c_\mathit{in},c_\mathit{out},c_\mathit{com},\mathit{eff})$ consisting of:
\begin{itemize}[noitemsep,topsep=0pt]
\item an alarm model $(c_\mathit{in},c_\mathit{com},\mathit{eff})$;

\item a function $c_\mathit{out}\in\mathcal{E}^+\times
2^{\mathcal{E}^+}\rightarrow \mathbb{R}^+_0$ to model the \emph{cost of
undesired \underline{out}come};
\end{itemize}
\end{definition}

\noindent
Next, we illustrate the above model with exemplary application scenarios.

\begin{exampleCont}[\ref{ex:unemployment}]
	While receiving unemployment benefits at UWV, several customers omit to inform
	UWV about the fact that they had found a job and, thus, kept receiving benefits
	that they were not entitled to. If the salary of the new job is lower than the
	salary of the previous job, then the customer is still entitled to a benefit
	proportional to the difference in salary. In fact, customers often mistakenly
	declare the wrong salary for their new job and are then expected to return a
	certain amount of received benefits.
	In practice, however, this rarely happens.
	UWV would therefore benefit from an alarm system that informs about customers
	who are likely to be receiving unentitled benefits.
	The \emph{cost of an intervention} stems from the hourly rate of UWV's
	employees; investigations using different IT systems and verifications with the old and new employer of a customer take a significant
	amount of time. The \emph{cost of undesired outcome} is the monetary value of
	the benefits that the customer received unlawfully.

Let $\mathit{unt}(\sigma)$ denote the amount of unentitled benefits received in a case corresponding to trace $\sigma$.
Based on discussions with UWV, we designed the following cost model.
\setlist[description]{font=\normalfont}
\begin{compactdesc}
  \item[\normalfont{Cost of intervention.}] For the intervention, an employee
  needs to check
  if the customer is indeed receiving unentitled benefits and, if so, fill in
  the forms for stopping the payments.
 Thus, the cost of intervention is the total labour cost of the employee for the amount of time they spend executing an intervention.
  \item[\normalfont{Cost of undesired outcome.}] The amount of unentitled
  benefits that the customer would obtain without stopping the payments.
  \item[\normalfont{Cost of compensation.}] The social security institution
  works in a situation of monopoly, which means that the customer cannot be
  lost because of moving to a competitor, i.e., there is no cost of
  compensation: $c_\mathit{com}(\sigma,L)=0$.
  \item[\normalfont{Mitigation effectiveness.}] The proportion of unentitled
  benefits that will not be paid thanks to the intervention.
  \end{compactdesc}
\end{exampleCont}

\noindent
Due to the lack of competition for public administration, the above example
does not include compensation costs. The following example from finance,
however, illustrates the importance of being able to
incorporate the cost of compensation.

\begin{example}[Financial Institute]\label{ex:bank}
  Suppose that the customers of a financial institute use their credit cards
  to make payments online. Each such transaction is associated with a risk of
  fraud, e.g., through a stolen or cloned card. In this scenario, an alarm
  system shall determine whether the card needs to be blocked,
  due to a high risk of fraud. However, superfluous blocking of the card causes
  discomfort to the customer, who may then switch to a different financial
  institute.

  \setlist[description]{font=\normalfont}
  \begin{compactdesc}
    \item[\normalfont{Cost of intervention.}] The card is automatically
    blocked by the system. Therefore, the intervention costs
    are limited to the costs for sending a new
    credit card to the customer by mail.
    \item[\normalfont{Cost of undesired outcome.}] The total amount of money
    related to
    fraudulent transactions that the bank would need to reimburse to the
    legitimate customer.%
    \item[\normalfont{Cost of compensation.}] It is defined as the expected
    asset of the
    customer that would be lost, namely the actual asset multiplied by a
    certain probability $p \in [0,1]$, which is the fraction of customers who
    left the institute within a short time after the card was wrongly
    blocked.
    Denoting the asset value of a customer
    (the amount of the investment portfolio, the account balance,
    etc.) with $\mathit{asset}(\sigma)$, the
    cost of compensation is estimated as:
    $c_\mathit{com}=p\cdot\mathit{asset}(\sigma)$.
    \item[\normalfont{Mitigation effectiveness.}] The proportion of the total
    amount of
    money related to fraudulent transactions that does not need to be
    reimbursed
    by blocking the credit card after that $k$ events have been executed.
  \end{compactdesc}
\end{example}

\subsection{Multi-Alarm Cost Model}
\label{sec:multicosts}
The above formalization of the cost model assumes that an alarm system supports
a single type of intervention. For some processes, however, multiple types of
alarms exist and represent alternative options to interfere in the execution of
a case. Consider Example~\ref{ex:bank}, the scenario of blocking a credit card
to prevent fraud. An alternative intervention is to call the credit card owner
by phone to verify the suspicious transaction. A phone call certainly has a
higher cost of intervention, in comparison to automatically blocking the credit
card. Yet, a phone call has a lower cost of compensation because, in case of
falsely suspected fraud, it prevents the inconvenience of unnecessarily
blocking a card.

The choice of which alarm to employ might depend on the likelihood of achieving
an undesired outcome, if no action is put in place.
For instance, for the financial institute, if fraud is assessed to be a
possibility, but not a near-certainty, then it is preferable to call the
customer. If it is assessed to be fraud with near-certainty, it is safer to
preventively block the card.

To handle scenarios with multiple alarms, the following definition generalizes
our earlier single-alarm cost model.
\begin{definition}[Multi-Alarm Cost Model]
\label{def:multicosts}
Let $A$ be a set of alarms and $C$ be the universe of alarm models.
An \emph{multi-alarm cost model} is a tuple $(A,{ma},c_\mathit{out})$
consisting of:
\begin{itemize}[noitemsep,topsep=0pt]
\item a function $ma \in A \rightarrow C$ mapping each alarm $a \in A$ to a
model $ma(a)$;
\item a function  $c_\mathit{out}\in\mathcal{E}^+\times
2^{\mathcal{E}^+}\rightarrow \mathbb{R}^+_0$ to model the \emph{cost of
undesired \underline{out}come};
\end{itemize}
\end{definition}
\noindent The above definition subsumes
Definition~\ref{def:singlecosts}: If only one alarm is present (i.e.,
$A=\{a\}$), there is only a one alarm model $ma(a)=c$.

\subsection{Alarm-Based Prescriptive Process Monitoring System}
\label{sec:costInstances}

An alarm-based prescriptive process monitoring system is driven by the outcome
of the cases. In the remainder, the outcome of the cases is represented by a
function $\mathit{out}: \mathcal{E}^+\rightarrow \{\true,\false\}$: given a
case identified by a trace $\sigma$, if the case has an undesired outcome,
$\mathit{out}(\sigma)=\true$; otherwise, $\mathit{out}(\sigma)=\false$. In
reality, during the execution of a case, its outcome is not yet known and needs
to be estimated based on past executions that are recorded in an event log $L
\subset \mathcal{E}^+$. The outcome estimator is a function $\widehat{out}_L:
\mathcal{E}^+\rightarrow[0,1]$ predicting the likelihood
$\widehat{out}_L(\sigma')$ that the outcome of a case that starts with prefix
$\sigma'$ is undesired. We define an alarm system as a function that
returns true or false depending on whether an alarm is raised based on the
predicted outcome.

\begin{definition}[Alarm-Based Prescriptive Process Monitoring System]

\label{def:alarm}
Given an event log $L \subset \mathcal{E}^+$, let $\widehat{out}_L$ be an outcome estimator built from $L$.
Let $A$ be the set of alarms that can be raised.
An \emph{alarm-based prescriptive process monitoring system} is a function
$\mathit{alarm}_{\widehat{out}_L}: \mathcal{E}^+\rightarrow A \cup
\{\bot\}$
Given a running case with current prefix $\sigma'$,
$\mathit{alarm}_{\widehat{out}_L}(\sigma')$ returns the alarm raised after
$\sigma'$, or $\bot$ if no alarm is raised.
\end{definition}
For simplicity, we omit subscript $L$ from $\widehat{out}_L$, and the entire
subscript $\widehat{out}_L$ from $\mathit{alarm}_{\widehat{out}_L}$ when it is
clear from the context. An alarm system can raise an alarm at most once per
case, since we assume that already the first alarm triggers an intervention by
the stakeholders.

\begin{table}[tb]
	\centering
\begin{footnotesize}
	\begin{tabular}{c|c|c}
		\toprule
		& undesired outcome & desired outcome \\
        & $\mathit{out}(\sigma)=\true$ & $\mathit{out}(\sigma)=\false$ \\
		\midrule
		alarm raised & $c_\mathit{in}(i_\sigma,\sigma,L) + (1-\mathit{eff}(i_\sigma,\sigma,L)) c_\mathit{out}(\sigma,L)$ & $c_\mathit{in}(i_\sigma,\sigma,L) + c_\mathit{com}(\sigma,L)$ \\
		alarm not raised & $c_\mathit{out}(\sigma,L)$ & $0$ \\
		\bottomrule
	\end{tabular}
\end{footnotesize}
\caption{Cost of a case with trace $\sigma$ based on its outcome and whether an
alarm was raised. If the alarm is raised, $i_\sigma$ indicates the index of
$\sigma$ when the alarm occurred.
}
\label{table:cost_matrix_business_costs}
\end{table}

\tablename~\ref{table:cost_matrix_business_costs} illustrates how the cost of a
case is determined based on a cost model on the basis of a multi-alarm cost
model $(A,{ma},c_\mathit{out})$. In the table, $i_\sigma$ indicates the index
of the event in $\sigma$ when the alarm was raised, namely the smallest $i \in
[1,|\sigma|]$ such that
$\mathit{alarm}(\mathit{hd}^{i_\sigma}(\sigma))\neq\bot$; also,
$ma(\mathit{alarm}(\mathit{hd}^{i_\sigma}(\sigma)))=(c_{in},c_{com},\mathit{eff})$.

The above definition provides a framework for alarm-based prescriptive process
monitoring systems. Next, we turn to the problem of tuning a prescriptive process monitoring system in order to optimize its net cost.

\section{Alarm Systems and Empirical Thresholding}
\label{sec:approach}
This section introduces four types of alarm systems. In
Section~\ref{sec:approach_basic}, we introduce a basic mechanism based on
empirical thresholding. In Section~\ref{sec:approach_delay},
we enhance this approach with the idea of delaying alarms. An enhancement based
on prefix-length-dependent thresholds is introduced in
Section~\ref{sec:approach_prefix}. Finally, multiple
possible alarms are handled in Section~\ref{sec:approach_multi}.

\subsection{Basic Alarm System}
\label{sec:approach_basic}

We first consider a basic alarm system, in which there is only one type of
alarm $a$, which is triggered as soon as the estimated probability of an
undesired outcome exceeds a given threshold $\tau$.
Given this simple alarm system, we aim at finding an optimal value for the alarming threshold ${\tau}$, which minimizes the cost on a log $L_{\mathit{thres}}$ comprising historical
traces such that $L_{\mathit{thres}}\cap L_{\mathit{train}}=\emptyset$
with respect to a given probability estimator
$\widehat{\mathit{out}}_{L_\mathit{train}}$ and alarm model $\mathit{ma(a)}$.

The total cost of a mechanism to raise alarms $\mathit{alarm}$ on a log $L$ is
defined as $\mathit{cost}(L,\mathit{ma(a)},\mathit{alarm})=\Sigma_{\sigma\in
L}\mathit{cost}(\sigma,L,\mathit{ma(a)},\mathit{alarm})$. Based thereon, we
define an optimal threshold as ${\tau} = \arg\min_{\tau'\in[0,1]} \mathit{cost}
(L_\mathit{thres},\mathit{ma(a)}, \mathit{alarm}_{\tau'})$. Optimizing a
threshold
$\tau$ on a separate thresholding set is called \emph{empirical
thresholding}~\cite{sheng2006thresholding}. The search for such a threshold
${\tau}$ wrt.\ a specified alarm model $ma$ and log
$L_{\mathit{thres}}$ is done through any hyperparameter optimization
technique, such as Tree-structured Parzen Estimator (TPE)
optimization~\cite{bergstra2011algorithms}. The resulting approach
is a form of cost-sensitive learning, since the value
${\tau}$ depends on how the alarm model $\mathit{ma(a)}$ is specified.

\subsection{Delayed Firing System}
\label{sec:approach_delay}

The basic alarm system introduced above fires an alarm as soon as the
probability of an undesired outcome is higher than the threshold $ \tau $. This
may lead to firing an alarm too soon. Consider a scenario in which,
after observing event $e_i$, the probability is above $ \tau $,
whereas it drops below $
\tau $ and stays below $
\tau $ after the subsequent event $e_{i+1}$ has been recorded.
The basic alarm system would fire the alarm even if the probability of an undesired outcome is above $ \tau $ for one single event and then drops below $ \tau $ in subsequent events in a case.

An alternative alarming system (herein called \emph{delayed firing}) is to
fire an alarm only if the probability of a negative outcome remains above
threshold $ \tau $ for $\kappa$ consecutive events, where $\kappa$ is the
\emph{firing delay}. One would expect this delayed firing system to be more
robust to instabilities in the predictive model (e.g.,\ robust to high levels of
variations in the probability calculated by the predictive model for
consecutive events in a trace).

When building a delayed alarming mechanism, we need to consider both, the
firing delay $\kappa$ as well as the threshold $\tau$, for the hyperparameter optimization.
 Note that the basic alarm system is a special case of the delayed firing system with a firing delay
of $\kappa = 1$.

\subsection{Prefix-length-dependent Threshold System}
\label{sec:approach_prefix}

To cope with scenarios with dynamic costs, we propose to not rely only on a
single global alarming threshold ${\tau}$, but to use different thresholds
depending on the length of the prefix. That is, a
separate threshold ${\tau}_k$ is optimized for each prefix length $k$ or, more
generally, thresholds ${\tau}_{a \rightarrow b}$ are optimized for certain
intervals of prefix lengths $[a, b]$ for $a,b \in \mathbb{N}$. For instance,
${\tau}_{1 \rightarrow
4}$ denotes the optimal threshold for prefix lengths 1 to 4, while
${\tau}_{5 \rightarrow \infty}$ denotes the optimal threshold from
length 5 to the end of the trace. To divide the whole range of prefix
lengths into $n$ non-overlapping intervals, we define $n$ \emph{splitting
prefixes} $ \rho_i, 1 \le i \le n $, where $\rho_i$ refers to the start point
of interval $i$. For instance, in the previous example, $\rho_1=1, \rho_2=5$. A
single global threshold and prefix-length-based thresholds can be seen as
special cases of interval-based thresholds.

The splitting prefixes can either be user-defined or optimized over $L_{\mathit{train}}$.
In a system with multiple thresholds, all the thresholds can be optimized
simultaneously. It is also possible to treat the splitting prefixes for prefix
intervals as hyperparameters and optimize them over the event log $L$ together
with the thresholds. Prefix-length dependent thresholds ${\tau}_{a \rightarrow
b}$ may be combined with a fire delay $\epsilon$ to build an alarm system.
Then, the fire delay $\epsilon$  and the prefix-length dependent thresholds
${\tau}_{a \rightarrow b}$ are trained at the same time.

\subsection{Multi-alarm systems}
\label{sec:approach_multi}

To optimize the decision of firing an alarm in a setting with multiple possible alarms $A$
we consider each alarm, as well as the option to not firing an alarm, as a potential
decisions. Therefore we formulate the problem similar to a multi-class classification task~\cite{tsoumakas2007multi},
with $|A \cup \{ \bot\} | $ different classes: one class
for no alarm and one class per alarm type. This might look like a traditional multi-class classification task,
however it is not. Because in such a task a perfect system would assign multiple labels and therefore we would also use a training set with
multiple labels. In our scenario a perfect system would only fire the cheapest alarm, with the lowest cost of intervention and/or mitigation effect,
for all cases that will end in an undesired outcome. For all other cases no alarm would be fired.
In conclusion a perfect solutions would not require multiple alarms and the training set would only consist of two labels.
Therefore, our can not be solved with a lot of multi-class classification techniques like one-vs-all approaches.

However, one way of solving a multi-class classification problem is to break it
into several binary classification problems (i.e., the \emph{one-vs-one}
approach)~\cite{allwein2000reducing}. Then, all classes are tested
against each other with a binary classification algorithm. The set of binary
decisions leads to a number of votes, that are assigned to the classes that won
the binary classifications. The input data is ascribed the class with the most
votes. Our optimization problem can modelled simultaneously like an  \emph{one-vs-one} approach.

Optimizing the alarms against the class of not firing an alarm is done by
applying the basic model to each alarm independently. In a scenario with two
possible alarms this gives us the thresholds $\tau_{False-vs.-a_1}$ and
$\tau_{False-vs.-a_2}$. All alarms are then trained against each other with
empirical thresholding. In our example, that implies training the threshold
$\tau_{a_1-vs.-a_2}$. However, for training $\tau_{a_1-vs.-a_2}$, we use only
the prefixes $\sigma$ that have a higher likelihood probability than both
thresholds for these alarms (e.g.,
$\widehat{\mathit{out}}_{L_\mathit{train}}(\mathit{hd}^k(\sigma)) >
\tau_{False-vs.-a_1}$ and
$\widehat{\mathit{out}}_{L_\mathit{train}}(\mathit{hd}^k(\sigma)) >
\tau_{False-vs.-a_2}$). While this might seem contrary to the one-vs-one
approach (the threshold is not trained over the whole dataset), it
is necessary for the following reasons.
First, the question of `should we
fire an alarm?' shall be separated from the question of `which alarm to fire?'.
The decision on the alarm type is taken only over the subset of prefixes where
alarming was found necessary. Training the threshold over all prefixes would
fit the threshold to irrelevant prefixes. Also, it is more important to
fire any alarm for a case with an undesired outcome than to choose the best
alarm.
Since we train the thresholds to decide between alarms in a hierarchical order, first the alarm vs. no alarm thresholds and then the alarms against each other, we call this approach \emph{hierarchical thresholding}.

\begin{figure}[hbtp]
\centering
\includegraphics[width=.75\textwidth]{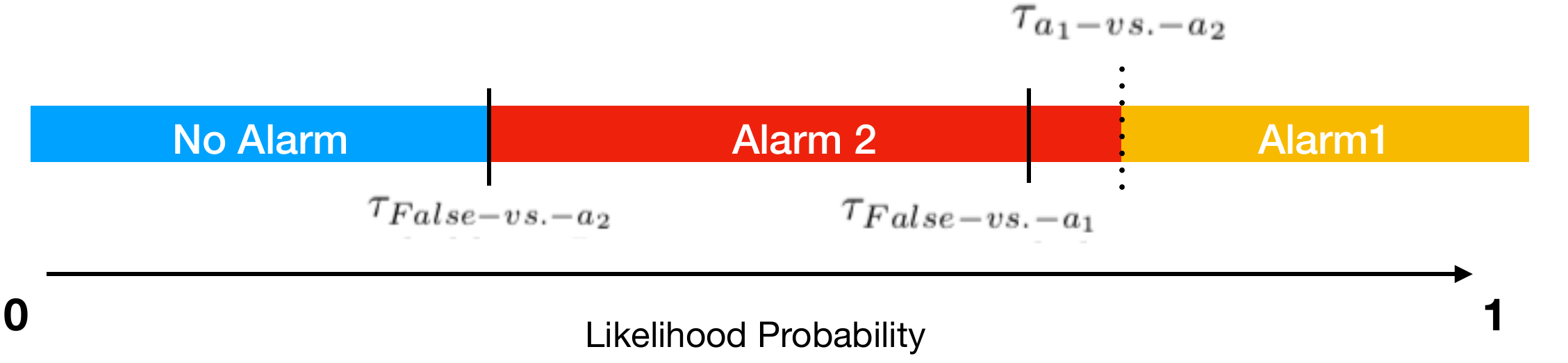}
\caption{Likelihood probability in hierarchical thresholding.}
\label{fig:multi_alarm_hierachical_decision}
\end{figure}

Figure~\ref{fig:multi_alarm_hierachical_decision} illustrates the general idea
for a system with two alarm types.
It represents the probability of an undesired outcome on the
horizontal axis. The three aforementioned thresholds are visualized as vertical
lines. Different colors illustrate the slides of the likelihood probability
that will lead to different actions by the alarm system: firing no alarm, or
firing one of the two alarms $a_1,a_2$.

\section{Evaluation}
\label{sec:evaluation}

In this section, we report on an experimental evaluation of the proposed
framework. Specifically, our evaluation addresses the following research
questions related to the overall effectiveness of alarm-based prescriptive
process monitoring:

\setlist[enumerate]{nosep}
\begin{enumerate}[label=RQ\arabic*,leftmargin=*]
\item \emph{Does empirical thresholding identify thresholds that consistently
reduce the average processing cost for different alarm model
configurations?}
\item \emph{Does the alarm system consistently yield a benefit over different
values
of the mitigation effectiveness?}
\item \emph{Does the alarm system consistently yield a benefit over different
values of the cost of compensation?}
\end{enumerate}
Moreover, we explore in detail the design choices involved when deciding on
when to fire an alarm, going beyond simple empirical thresholding.

\medskip
\noindent
\emph{Is there a reduction in the average processing cost per case:}
\begin{enumerate}[label=RQ\arabic*,leftmargin=*]
  \setcounter{enumi}{3}
  \item \emph{When training a parameter for the minimum number of events that
  exceed the threshold?}
  \item \emph{When using more than one threshold interval?}
  \item \emph{When increasing the number of prefix-length-dependent thresholds?}
  \item \emph{When combining prefix-length dependent thresholds with a firing
  delay,
compared to the a single-alarm system?}
\end{enumerate}
Finally, we turn to a comparison of systems that feature a single alarm only
and those that support multiple alarms.
\begin{enumerate}[label=RQ\arabic*,leftmargin=*]
  \setcounter{enumi}{7}
  \item \emph{Is there a reduction in the average processing cost per case when
  using multiple alarms compared to the best system with only one alarm?}
\end{enumerate}
In the remainder, we first discuss the real-world datasets used in our
evaluation (\autoref{sec:eval_datasets}), before turning to the experimental
setup (\autoref{sec:eval_setup}). We then report on our evaluation results in
detail and close with an overview of our answers to the above research questions
(\autoref{sec:eval_results}).

\subsection{Datasets}
\label{sec:eval_datasets}

We use the following real-world datasets to evaluate the alarm system:
\begin{compactdesc}
  \item[BPIC2017.]  This log contains traces of a loan application
  process in a Dutch
  bank.\footnote{\url{https://doi.org/10.4121/uuid:5f3067df-f10b-45da-b98b-86ae4c7a310b}}
  It was split into two sub-logs, denoted with
  \emph{bpic2017\_refused} and \emph{bpic2017\_cancelled}. In the first one,
  the undesired cases refer to the process executions in which the applicant
  has refused the final offer(s) by the financial institution. In the
  second one, the undesired cases consist of those cases where the financial
  institution has cancelled the offer(s).
  \item[Road traffic fines.] This log originates from a Italian
  police unit and relates to a process to collect traffic
  fines.\footnote{\url{https://doi.org/10.4121/uuid:270fd440-1057-4fb9-89a9-b699b47990f5}}
  The desired outcome is that a fine is paid, while in the undesired
  cases the fine needs to be sent for credit collection.
  \item[Unemployment.] This event log corresponds to the \emph{Unemployment
  Benefits} process run by the UWV in the Netherlands, introduced already as
  \autoref{ex:unemployment} in \autoref{sec:framework}. Due to privacy
  constraints, this event log is not publicly available.
  The undesired outcome of the process is that a resident will
  receive more benefits than entitled, causing the need for a reclamation.

\end{compactdesc}
\autoref{table:dataset_stats} describes the characteristics of the event logs
used. These logs cover diverse evaluation settings, along several dimensions.
The classes are well balanced in \emph{bpic2017\_cancelled} and
\emph{traffic\_fines}, while the undesired outcome is more rare in
\emph{unemployment} and \emph{bpic2017\_refused}. In \emph{traffic\_fines}, the
traces are very short, while in the other datasets the traces are generally
longer.

For each event log, we use all available data attributes as input to the classifier. Additionally, we extract the \emph{event number}, i.e., the index of the event in the given case, the \emph{hour, weekday, month, time since case start}, and \emph{time since last event}.
Infrequent values of categorical attributes (occurring less than 10 times in
the log) are replaced with value `other', to avoid a massive blow-up of the
considered number of dimensions. Missing attributes are imputed with the
respective most recent (preceding) value of that attribute in the same trace
when available, otherwise with zero.
Traces are cut before the labeling of the case becomes trivially known and are truncated at the 90th percentile of all case lengths to avoid bias from very long traces.

\begin{table}[t]
\caption{Dataset statistics}
\label{table:dataset_stats}
\vspace{-1.5em}
\footnotesize
\begin{center}
	\begin{tabular}{@{}lrrrrrr@{}}
	\toprule
  & \multicolumn{1}{c}{\#} & \multicolumn{1}{c}{class} &
  \multicolumn{1}{c}{min} & \multicolumn{1}{c}{med} &
  \multicolumn{1}{c}{(trunc.) max} & \multicolumn{1}{c}{\#} \\
dataset name &  \multicolumn{1}{c}{traces} & \multicolumn{1}{c}{ratio} &
\multicolumn{1}{c}{length} & \multicolumn{1}{c}{length} &
\multicolumn{1}{c}{length} &  \multicolumn{1}{c}{events} \\ \midrule
bpic2017\_refused & 31\,413 & 0.12 & 10 & 35 & 60 & 1\,153\,398 \\
bpic2017\_cancelled & 31\,413 & 0.47 & 10 & 35 & 60 & 1\,153\,398 \\
traffic\_fines & 129\,615 & 0.46 & 2 & 4 & 5 & 445\,959 \\
unemployment & 34\,627 & 0.20 & 1 & 21 & 79 & 1\,010\,450 \\
	\bottomrule
	\end{tabular}
\end{center}
\vspace{-0.2cm}
\end{table}

\subsection{Experimental Setup}
\label{sec:eval_setup}

We split the aforementioned datasets temporally, as
follows. We order cases by their start time and randomly select 80\% of the
first 80\% of the cases (i.e., 64\% of the total) for $L_\mathit{train}$;
20\% of the first 80\% of the cases (i.e., 16\% of the total) for
$L_\mathit{thres}$; and use the remaining 20\% as the test set
$L_\mathit{test}$. The events in cases in $L_\mathit{train}$ and
$L_\mathit{thres}$ that overlap in time with $L_\mathit{test}$ are discarded in
order to not use any information that would not be available yet in a real
setting.

Based on $L_\mathit{train}$, we build the classifier $\widehat{\mathit{out}}$
using
random
forest (RF) and gradient boosted trees (GBT).
Both algorithms have been shown to work well on a variety of classification
tasks~\cite{fernandez2014we,olson2017data}.
Features for a given prefix
are obtained using the aggregation encoding~\cite{de2016general}, which is
known to be effective for logs~\cite{teinemaa2017outcome}.

The configuration of the alarming mechanism depends on the setup
chosen for a specific research question. For RQ1 to RQ3, we rely on the basic
model, introduced in
\autoref{sec:approach_basic}, and determine an optimal alarming
threshold $\overline{\tau}$ based on $L_\mathit{thres}$. We employ
Tree-structured Parzen Estimator (TPE)
optimization~\cite{bergstra2011algorithms} with 3-fold cross validation. The
resulting alarm system is then compared against several baselines: First, we
compare with the \emph{as-is} situation, in which alarms are never raised.
Second, define a baseline with $\tau = 0$, which enables us to compare with the
situation where alarms are always raised directly at the start of a case.
Finally, setting $\tau = 0.5$, we consider a comparison with the
cost-insensitive scenario that simply raises alarms when an undesired outcome
is expected.

To answer RQ4 to RQ7, we consider the advanced models introduced in
\autoref{sec:approach_delay} and
\autoref{sec:approach_prefix}. Again, the parameters are
optimized using TPE. For RQ4, this includes the threshold
$\overline{\tau}$ and the firing delay $\kappa \in \{1,\ldots,7\}$. For RQ5,
we train two prefix-length-interval-based thresholds $\overline{\tau_{1
\rightarrow \rho}}$,
$\overline{\tau_{\rho \rightarrow \infty}}$ and optimize them together with
parameter $\rho$. For RQ6, we
define three different systems with 1 to 3 prefix length intervals.
We rely on user-defined intervals for the prefix length, as
follows. We set the length for each interval, except for the last, to
one prefix and put all intervals after each other staring at prefix length 1.
This results in three systems with
$ \{ \overline{\tau_{1
\rightarrow \infty}}  \} $, $ \{ \overline{\tau_{1 \rightarrow 2}},
\overline{\tau_{2 \rightarrow \infty}}\} $, and $ \{ \overline{\tau_{1
\rightarrow 2}}, \overline{\tau_{2 \rightarrow 3}}, \overline{\tau_{3
\rightarrow \infty}}\}$.
Finally, to test RQ7, we build an alarming mechanism by training thresholds
$\overline{\tau_{1 \rightarrow \rho}}$, $\overline{\tau_{\rho \rightarrow
\infty}}$, the interval-section point $ \rho$, and the firing delay $ \kappa
\in \{1,\ldots,7\}$.

To explore RQ8, we derive a multi-alarm system that optimizes the alarming
threshold $\overline{\tau}$, using TPE, for two different alarm types
independently (\autoref{sec:approach_multi}). For these alarms, we multiply the
factors given in
\autoref{table:alarm_factors} with the cost of intervention $c_{in}$ and the
cost of intervention $c_{in}$.
We compare the resulting system against a baseline
that always uses one of the alarms, i.e., the one that is better in terms of
average processing cost per case.

\begin{table}[!htb]
  \footnotesize
  \caption{Factors for the different alarms, that are multiplied with the
    respective costs}
  \label{table:alarm_factors}
  \vspace{-1.7em}
  \begin{center}
    \begin{tabular}{@{}lcccc@{}}
      \toprule
      & factor $c_\mathit{in}$ & factor $c_\mathit{com}$  \\ \midrule
      Alarm 1 & 1  & 1  \\
      Alarm 2 & 1.2 & 0.5 \\
      \bottomrule
    \end{tabular}
  \end{center}
  \vspace{-0.6cm}
\end{table}

It is common in cost-sensitive learning to apply calibration techniques
to the resulting classifier~\cite{zadrozny2001learning}. Yet, we found that
calibration
using Platt scaling~\cite{platt1999probabilistic} does not
consistently improve the estimated likelihood of the undesired outcome on our
data and, thus, did not apply calibration.

Next, we turn to the alarm models used in our evaluation,
summarized in \autoref{table:cost_models_eval}. For RQ1, we vary the ratio
between the cost of the undesired
outcome $c_\mathit{out}$
and the cost of intervention $c_\mathit{in}$, keeping the cost of compensation
$c_\mathit{com}$ and
the mitigation effectiveness $\mathit{eff}$ unchanged. The same is done for
RQ2 and, in addition, the mitigation effectiveness
$\mathit{eff}$ is varied. For RQ3, we vary two ratios: the one between
$c_\mathit{out}$ and $c_\mathit{in}$, and the one between
$c_\mathit{in}$ and $c_\mathit{com}$.

In the experiments related to RQ4-RQ7, we consider three types of
configurations,
each corresponding to one row in \autoref{table:cost_models_eval}. We
vary the cost of interference $c_\mathit{in}$ and the cost of compensation
$c_\mathit{com}$ from values that render them
insignificant compared to $c_{out}$, to values that yield a significant impact.
Specifically, the first type
of alarm model, coined constant cost configurations, assigns
constant
costs over the whole trace. A second type, linear cost configurations,
assigns costs that increase with longer trace prefixes. The third type,
non-monotonic cost configurations, changes costs non-linearly
over the length of the trace.
Constants for this setting are introduced as
listed in \autoref{table:constants_nonmonotonic}. We assign values similar to
the minimum case length to constants used as a numerator and values
similar to the medium case length for constants used as a divisor.
This yields a non-monotonic
cost development over the trace length for most cases.

For RQ8, we largely follow the same setup. However, we define our test
scenarios such that the alarm that is more expensive for true positives is at
least 10\% cheaper for false positives compared to the other alarm. Thereby,
both alarms show a certain difference in the
induced cost trade-off.

\begin{table}[t]
\vspace{-0.2cm}
\caption{Alarm model configurations}
\label{table:cost_models_eval}
\vspace{-1.5em}
\begin{center}
\resizebox{1\textwidth}{!}{
	\begin{tabular}{@{}lcccc@{}}
	\toprule
 & $c_\mathit{out}(\sigma,L)$ & $c_\mathit{in}(k,\sigma,L)$ &
 $c_\mathit{com}(\sigma,L)$ & $\mathit{eff}(k,\sigma,L)$ \\
 \midrule
RQ1 & $1, 2, 3, 5, 10, 20$ & $1$ & $0$ & $1 - k / |\sigma|$ \\
RQ2 & $1, 2, 3, 5, 10, 20$ & $1$ & $0$ & $0, 0.1, 0.2,\ldots, 1$ \\
RQ3 & $1, 2, 3, 5, 10, 20$ & $1$ & $\{0, 1/20, 1/10, 1/5, 1/2, 1, 2, 5, 10,
20\}$ & $1 - k / |\sigma|$ \\[1em]
         & $10$ & $1,2,3,4,5$ & $0,1,2,3,4,5, 10,15, 20$ & $1$\\
RQ4--RQ7 & $10$ & $\{1,2,3,4,5\} \cdot k / |\sigma|$ &
$0,1,2,3,4,5, 10, 20$ & $1 - k/ |\sigma|$ \\
         & $10$ & $\{1,2,3,4,5\}*(1-\frac{\min(a,k-1)}{b})$ & $\{0,1,2,3,4,5,
         10, 20\}  * (1-\frac{\min(k-1,c)}{d})$ & $1 -
         \min(\frac{\min(e,k-1)}{f})$\\[1em]
RQ8 & $10$ &  $1,2,3,4,5$  &  $1, 2, 3, 4, 5, 10, 20, 30, 40$ & $1$\\
	\bottomrule
	\end{tabular}
	}
\end{center}
\vspace{-0.5cm}
\end{table}

To evaluate the success of prescriptive process monitoring, we measure the
\emph{average cost per case} using the test set $L_\mathit{test}$ derived for
each dataset as discussed above. This cost shall be minimal.
Moreover, we measure the \emph{benefit} of the alarm system, i.e., the
reduction in the average cost of a case when using the alarm system compared to
the average cost when not using it.

Additionally, we use the f-score to test the accuracy of
our systems, in terms of how often they fire
an alarm correctly. It is calculated as the harmonic mean of precision and
recall and ranges from 0 (worst) to 1 (best).

The above experimental setup has been implemented in Python based on
Scikit-Learn and LightGBM, with the prototype
being publicly available
online.\footnote{\url{https://github.com/samadeusfp/alarmBasedPrescriptiveProcessMonitoring/}}

\begin{table}[]
\caption{Constants for non-monotonic cost configurations}
\label{table:constants_nonmonotonic}
\footnotesize
\vspace{-1.7em}
\begin{center}
\begin{tabular}{l l l l l l l}
\toprule
Dataset             & a  & b  & c  & d  & e  & f  \\ \midrule
bpic2017\_cancelled & 10 & 35 & 13 & 32 & 18 & 40 \\
bpic2017\_refused   & 8  & 33 & 15 & 34 & 20 & 35 \\
traffic\_fines      & 3  & 5  & 2  & 5  & 3  & 4  \\
\bottomrule
\end{tabular}
\end{center}
\vspace{-1.7em}
\end{table}

\subsection{Results}
\label{sec:eval_results}

\begin{figure}[t]
\vspace{-0.2cm}
\centering
\includegraphics[width=.85\textwidth]{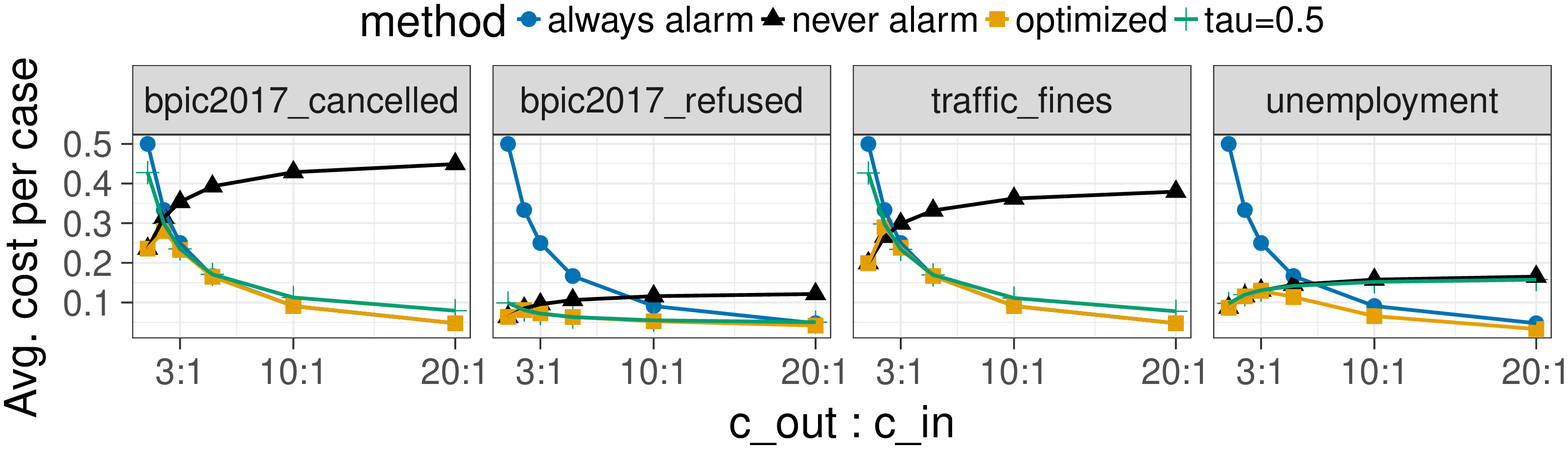}
\vspace{-.4cm}
\caption{Cost over different ratios of $c_\mathit{out}$ and $c_\mathit{in}$
(GBT)}
\label{fig:results_ratios}
\vspace{-0.2cm}
\end{figure}

We evaluate our basic model of an alarming system that is based on empirical
thresholding by exploring whether it consistently reduces the average
processing cost under different alarm models (RQ1).
\autoref{fig:results_ratios} shows the average cost per case when varying the
ratio of $c_\mathit{out}$ and $c_\mathit{in}$. We only present the results
obtained with GBT, which slightly outperform those with RF. When the ratio
between the two costs is balanced, the minimal cost is obtained
by never alarming.
When $c_\mathit{out} \gg c_\mathit{in}$, one shall always raise an alarm.
However, when $c_\mathit{out}$ is slightly higher than $c_\mathit{in}$, the
best strategy is to sometimes raise an alarm based on $\widehat{out}$.
We found that the optimized $\overline{\tau}$ always outperforms the baselines.
An exception is ratio 2:1 for the \emph{traffic\_fines} dataset, where never
alarming is slightly better. %

\begin{figure}[t]
\vspace{-0.2cm}
\centering
\includegraphics[width=.85\textwidth]{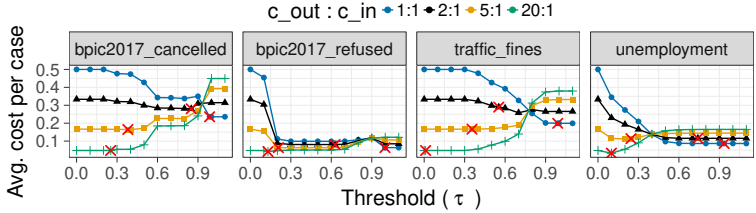}
\vspace{-.4cm}
\caption{Cost over different thresholds ($\overline{\tau}$ is marked with a red
cross)}
\label{fig:results_thresholds}
\vspace{-0.2cm}
\end{figure}

The impact of the threshold for firing an alarm is further explored in
\autoref{fig:results_thresholds}. The optimized threshold is marked with a red
cross and each line represents one particular cost ratio.
While the optimized threshold generally obtains minimal costs,
there sometimes exist multiple optimal thresholds for a given alarm model. For
instance, for the 5:1 ratio in
\emph{bpic2017\_cancelled}, all thresholds between 0 and 0.4 are cost-wise
equivalent. Hence, empirical thresholding consistently
finds a threshold that yields the lowest cost for a given log and cost
model configuration.

\begin{figure}[t]
\vspace{-0.2cm}
 \label{fig:results_heatmaps}
		\hspace{-0.25cm}
        \subfloat[Varying
        $\mathit{eff}$\label{fig:results_effectiveness_const_lgbm_selected}]{\includegraphics[width=0.515\linewidth]{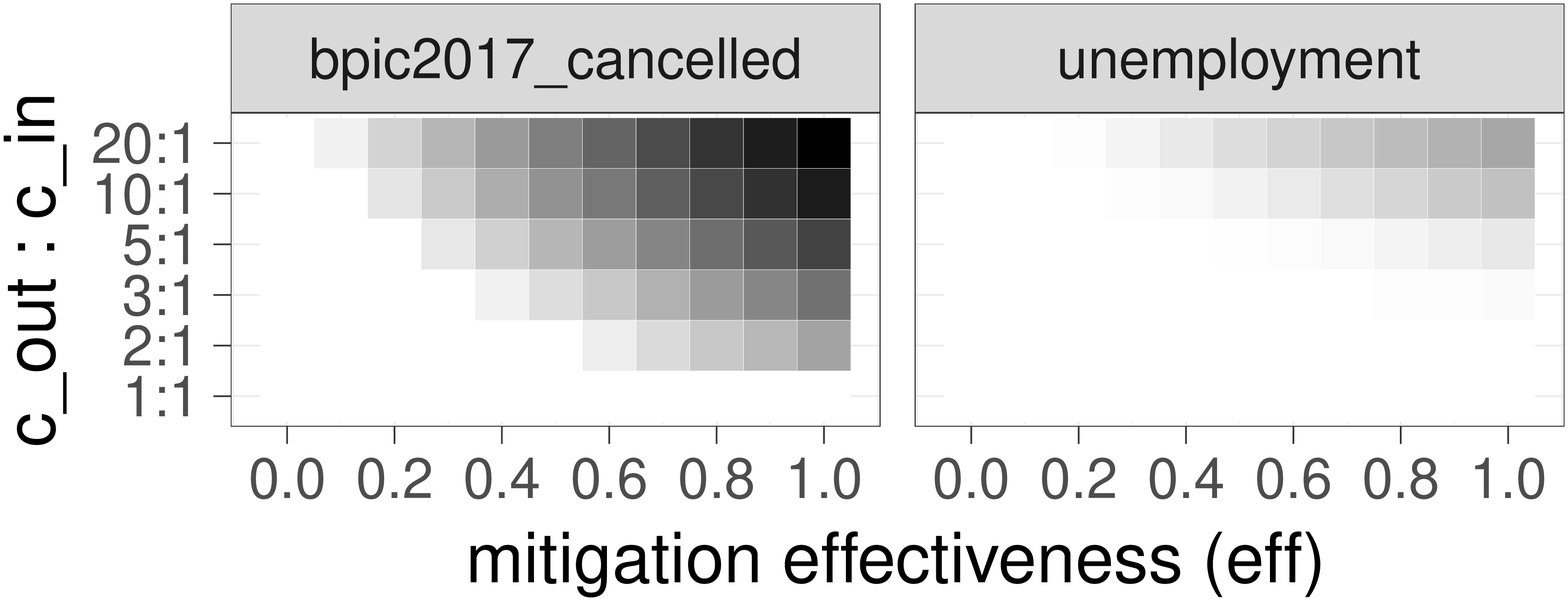}}
        \hspace{0.05cm}
        \subfloat[Varying
        $c_\mathit{com}$\label{fig:results_compensation_lgbm_selected}]{\includegraphics[width=0.5\linewidth]{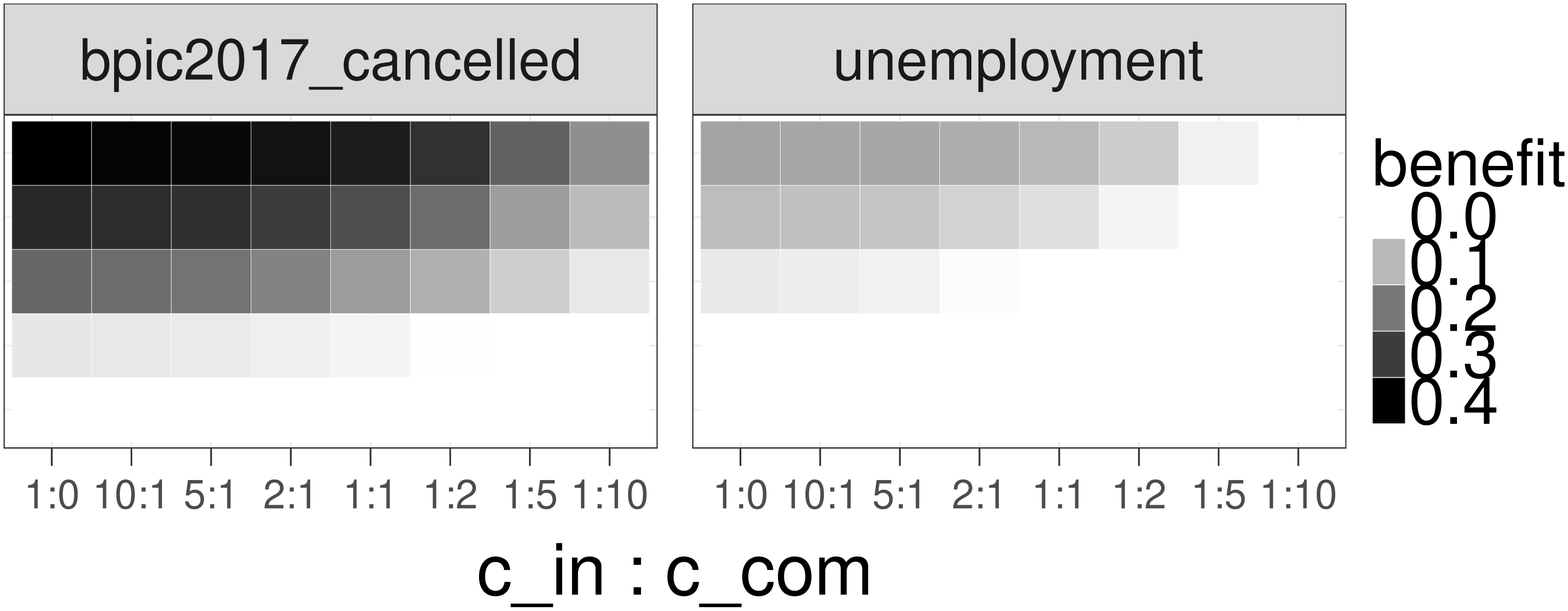}}
    \vspace{0.505\baselineskip}
\vspace{-.7cm}
    \caption{Benefit with different alarm model configurations
    }
    \vspace{-0.3cm}
\end{figure}

Turning to RQ2, we evaluate how the mitigation effectiveness influences the
results. Figure~\ref{fig:results_effectiveness_const_lgbm_selected} shows
the
benefit of having an alarm system compared to not having it for different
(constant) effectiveness values. As the results are similar for logs with
similar class ratios, hereinafter, we show the results for
\emph{bpic2017\_cancelled} (balanced classes) and
\emph{unemployment} (imbalanced classes). As expected, the benefit increases
both with higher $\mathit{eff}$ and with higher $c_\mathit{out}:c_\mathit{in}$
ratios. For \emph{bpic2017\_cancelled}, the alarm system yields a benefit when
$c_\mathit{out}:c_\mathit{in}$ is high and $\mathit{eff} > 0$. Also, a benefit
is always obtained when $\mathit{eff} > 0.5$ and $c_\mathit{out} >
c_\mathit{in}$. In the case of the \emph{unemployment} dataset, the average
benefits are smaller, since there are fewer cases with undesired outcome and,
therefore, the number of cases where $c_\mathit{out}$ can be prevented by
alarming is lower. In this case, a benefit is obtained when both $\mathit{eff}$
and $c_\mathit{out}:c_\mathit{in}$ are high.
We conducted analogous experiments with a linear decay in effectiveness,
varying the maximum possible effectiveness (at the start of the case), which
confirmed that the observed patterns remain the same.
As such, we have confirmed empirically that an alarm system yields a benefit
over different values of mitigation effectiveness.

Next, we consider the influence of the cost of compensation (RQ3). As above,
the benefit of the alarm system is plotted in
Figure~\ref{fig:results_compensation_lgbm_selected} across different ratios of
$c_\mathit{out}:c_\mathit{in}$ and $c_\mathit{in}:c_\mathit{com}$. When the
cost of compensation $c_\mathit{com}$ is high, the benefit
decreases due to false alarms. For \emph{bpic2017\_cancelled}, a benefit is
obtained almost always, except when $c_\mathit{out}: c_\mathit{in}$ is low
(e.g., 2:1) and $c_\mathit{com}$ is high (i.e., higher than $c_\mathit{in}$).
For \emph{unemployment}, fewer
configurations are beneficial, e.g., when $c_\mathit{out}:c_\mathit{in} = 5:1$
and $c_\mathit{com}$ is smaller than $c_\mathit{in}$.
We conducted analogous experiments with a linearly increasing cost of
intervention, varying the maximum possible cost, which
confirmed the above trends.
In sum, we confirmed empirically that the alarm system achieves a benefit
if the cost of the undesired outcome is sufficiently higher than the cost
of the intervention and/or the cost of the intervention is sufficiently higher
than the cost of compensation.

\begin{figure}[t]
  \centering
  \includegraphics[width=.85\textwidth]{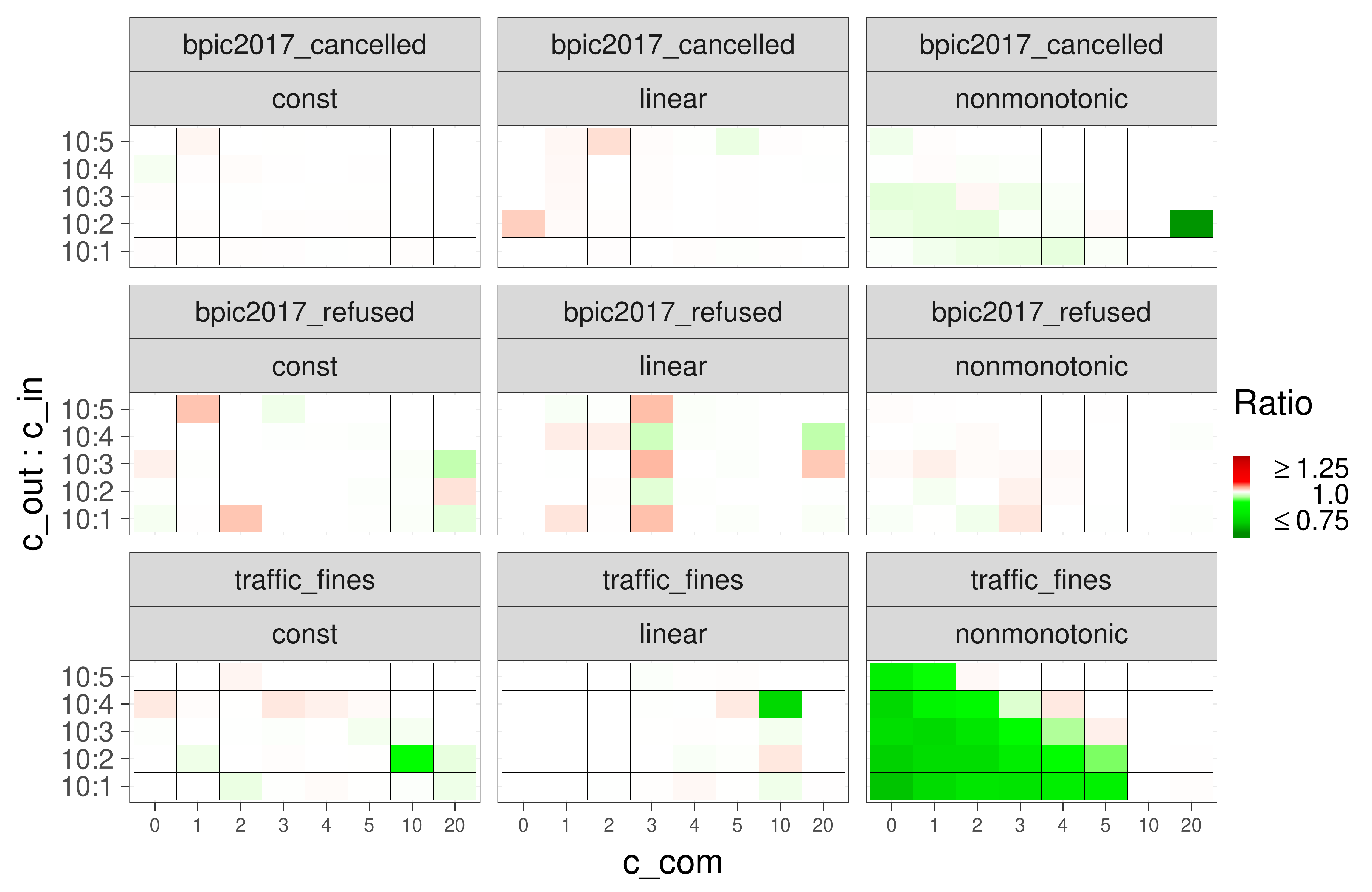}
\vspace{-.4cm}
  \caption{Ratio of the average processing costs for a system
  with and without firing delay}
  \label{fig:results_fire_delay_ratio}
\vspace{-.2cm}
\end{figure}

Next, we evaluate the design choices involved when
deciding on when to fire an alarm. We explore the reduction in
the average processing cost under a firing delay, the use of multiple
threshold intervals or prefix-length-dependent thresholds, and the combination
of the latter thresholds and a firing delay.

\begin{figure}[t]
  \centering
  \includegraphics[width=.85\textwidth]{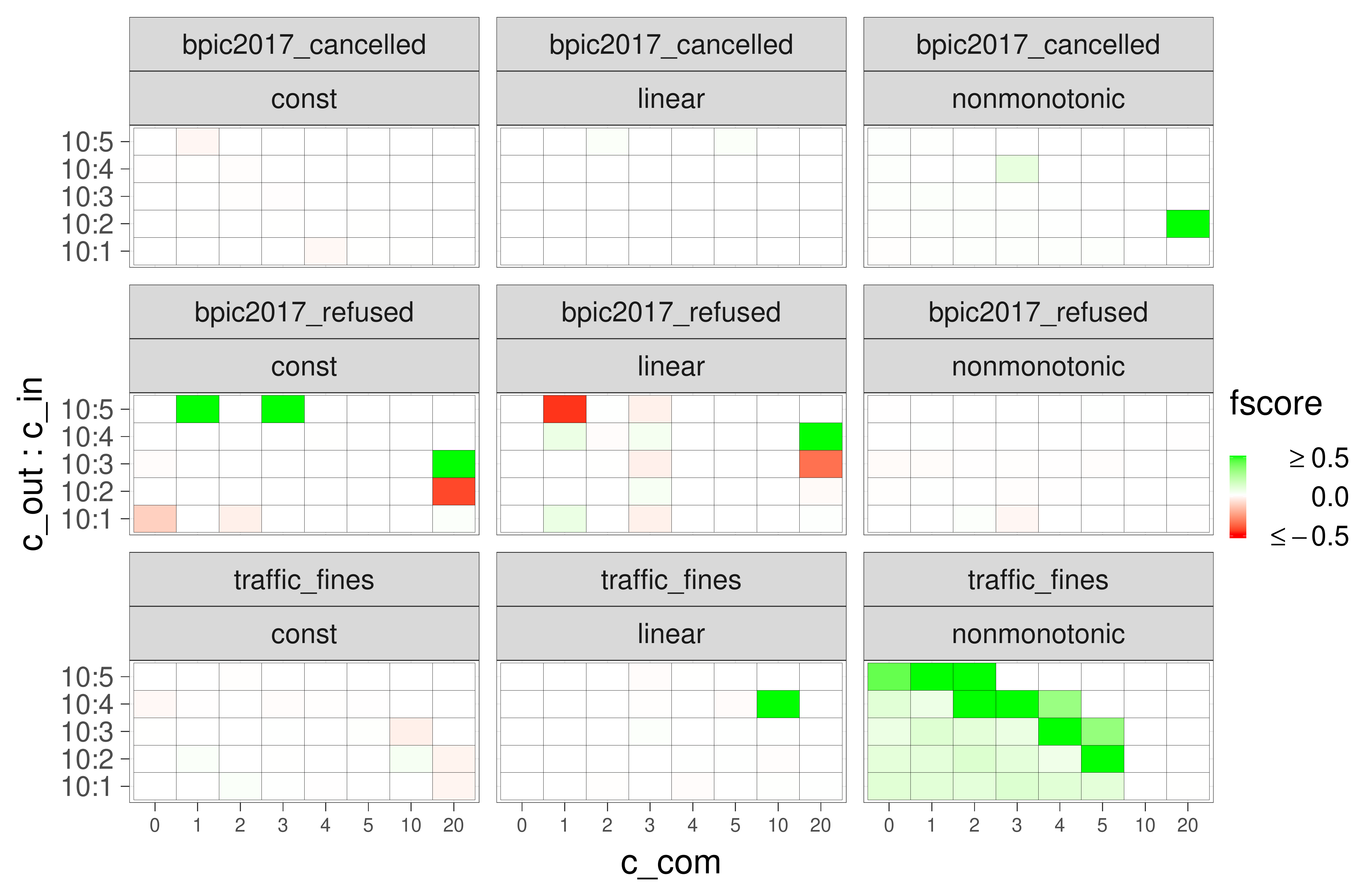}
  \vspace{-.4cm}
  \caption{F-score benefit for a system with and without firing
    delay}
  \label{fig:results_fire_delay_fscore}
  \vspace{-.4cm}
\end{figure}

To answer RQ4 on the effectiveness of a firing delay,
we calculated the ratio of the average cost per case
for the system with the firing delay as an additional hyperparameter, divided
by  the system without a firing delay.
\autoref{fig:results_fire_delay_ratio} shows nearly no cost
settings, in which the system that fires immediately outperforms the system
with the firing delay (red cells). Many times, the firing
delay reduces the costs (green cells), especially for the non-monotonic
scenarios. Sometimes
the costs are reduced by 20\% or more. However, in the vast majority of cases,
both systems produce the same results. This is due to the firing delay
parameter being set to 1 in many scenarios.
In addition to the ratio, we
also visualized the benefit in terms of f-score that the system with firing
delay delivers, see \autoref{fig:results_fire_delay_fscore}.
The results provide evidence for our hypothesis that waiting a certain number
of events before firing an alarm improves the classification of traces.
Overall, our results support a positive answer to RQ4: It is
possible to reduce the average processing cost per case with a firing delay
parameter.

\begin{figure}[h!]
\centering
\includegraphics[width=.85\textwidth]{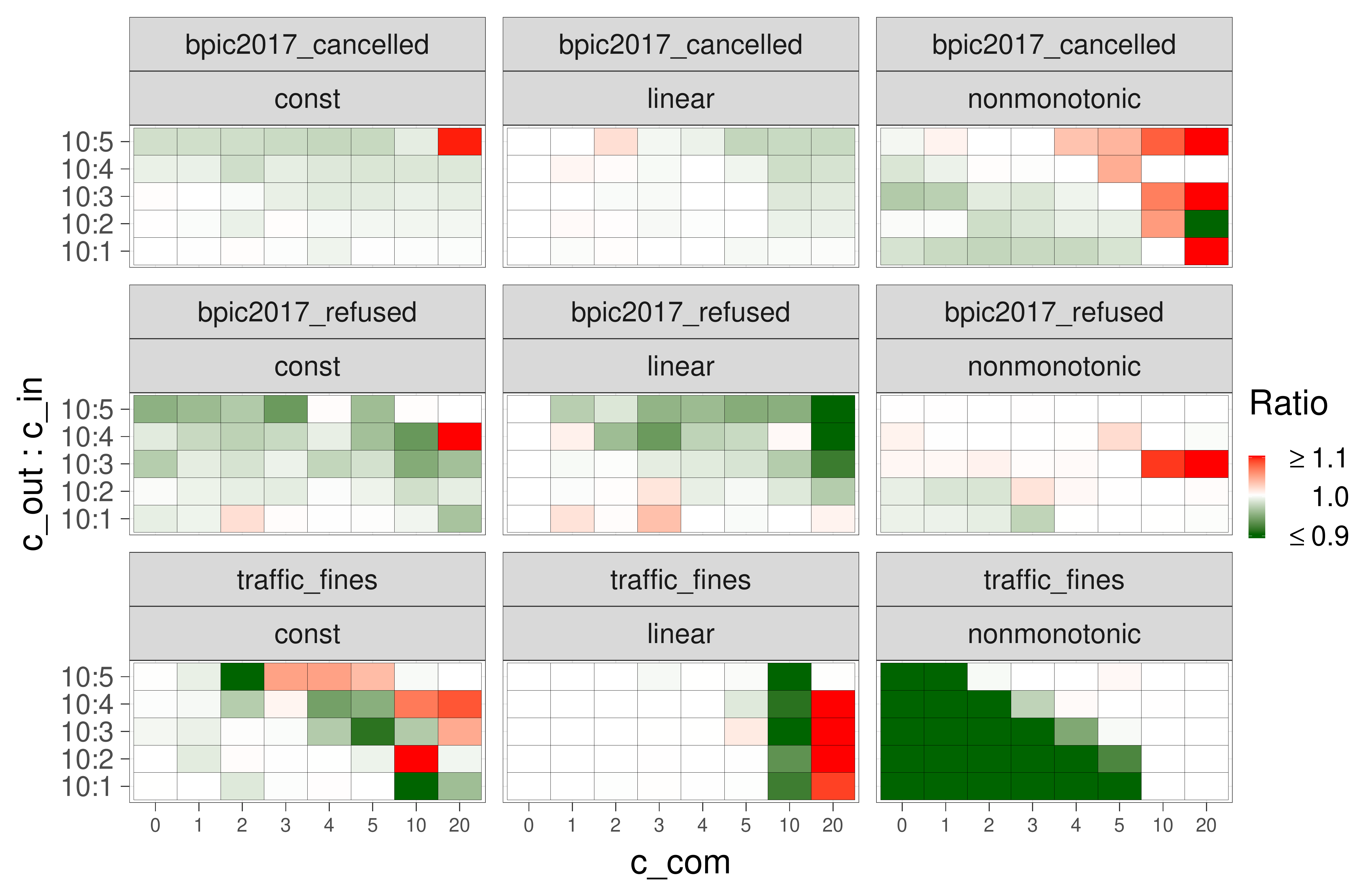}
\vspace{-.4cm}
\caption{Ratio of average processing costs for systems with a
single threshold or two thresholds}
\label{fig:results_heatmap_single_vs_prefix_ratio}
\vspace{-.5cm}
\end{figure}

To visualize relative benefits of multiple threshold intervals (RQ5), we
calculate the ratio of the average processing costs between our baseline, a
system based on the basic model, and our system with two threshold intervals.
In \autoref{fig:results_heatmap_single_vs_prefix_ratio}, we plot this ratio for
all used cost configurations. Green and red coloring represents an improvement
by using interval-based thresholds or by using a single global threshold,
respectively. Contrary to our expectation, in some scenarios the system with a
global
threshold outperforms a system with intervals. However, improvements are
observed for most scenarios.

Consistent improvements materialize for scenarios with non-monotonic
changing costs and the cost of compensation being lower than the cost of
undesired outcome. The improvements are the highest for the \textit{traffic
fines} dataset, that has the shortest traces. This leads also to shorter
intervals.
The smallest improvement in this group is seen in the \textit{bpic2017 refused}
dataset, which shows the lowest ratio of traces with an undesired outcome.
Due to this low class ratio, the potential room for improvement is also the
smallest, because for a smaller number of cases it is necessary to intervene.

Moreover, we observe
consistent improvements for the two \emph{bpic2017} datasets for linear
changing costs
and constant costs.

The \emph{traffic fines} dataset with linear costs is an interesting outlier.
It shows nearly no improvement for scenarios, in which the cost of compensation
is lower
than the cost of undesired outcome. If the cost of undesired outcome has the
same value as the cost of compensation the system with threshold intervals
performs significantly better than the system with a global threshold. If
the cost of compensation is higher than the cost of undesired outcome, the
system with a single global threshold is significantly better. We investigated
why the interval systems performs
worse than the single threshold system and
found out that, in the linear scenario, the single threshold system set the
threshold to such a high level, that the threshold is never reached and no
alarm is fired. However, this is not true for the system with two threshold
intervals, this system fire alarms in a variety of cases. This approach
underperforms here compared to not firing an alarm at all. The reason for this
seems to be overfitting.

\begin{figure}[t]
  \centering
  \includegraphics[width=.85\textwidth]{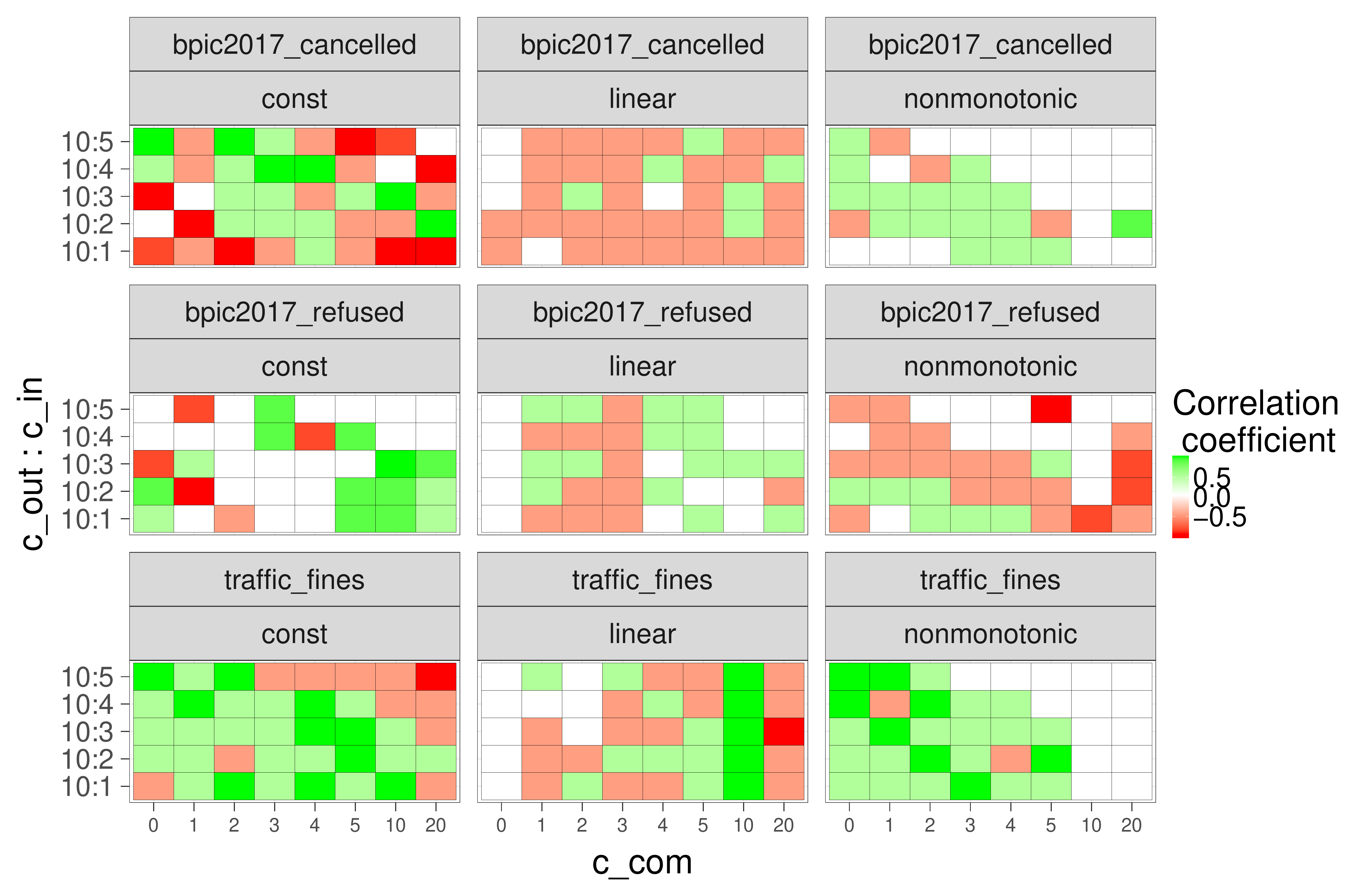}
  \vspace{-.4cm}
  \caption{Correlation coefficient between number of intervals and average cost
    per case}
  \label{fig:results_prefix_thresholds_corel}
  \vspace{-.3cm}
\end{figure}

\begin{figure}[h!]
  \centering
  \includegraphics[width=.85\textwidth]{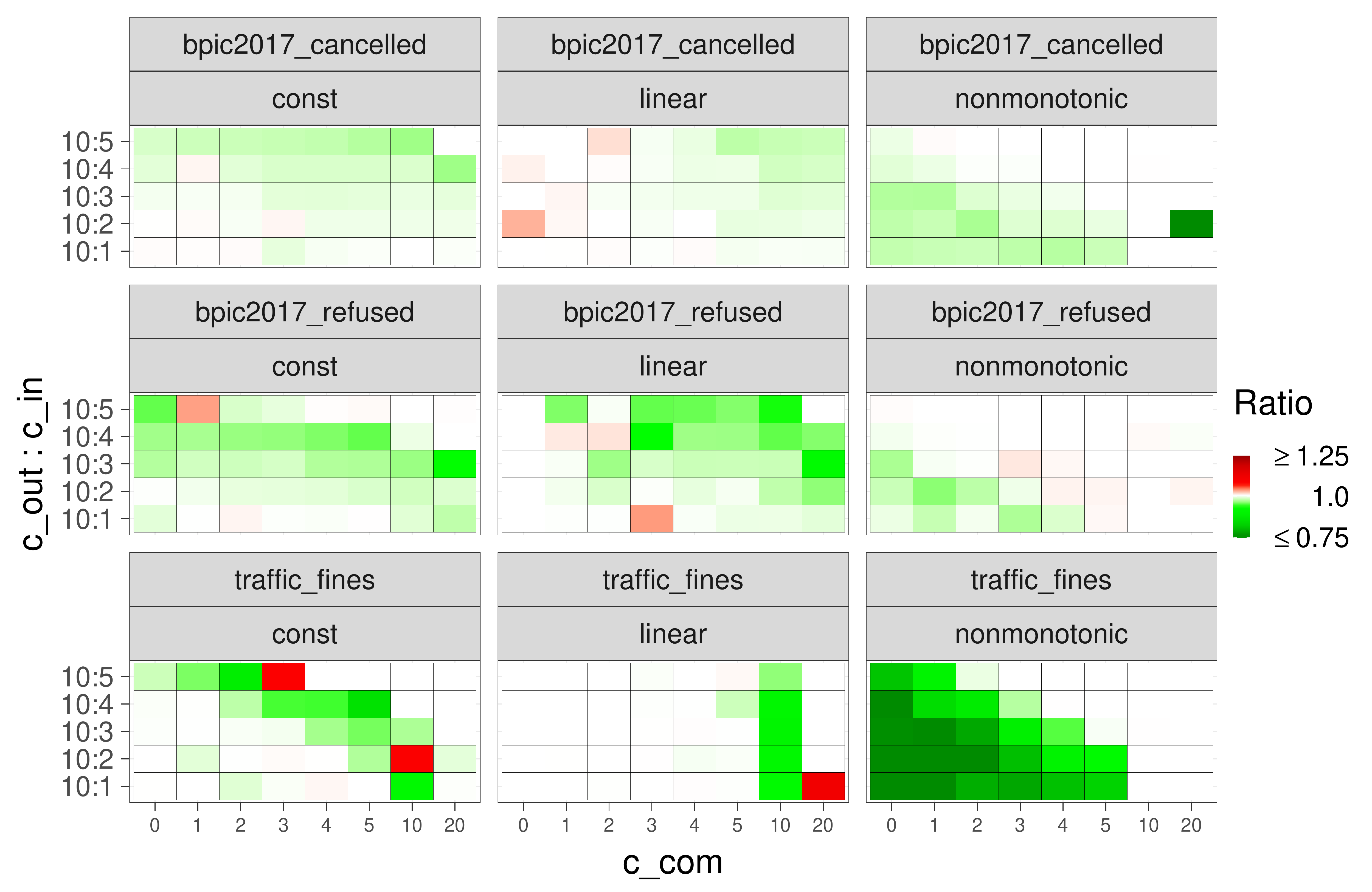}
  \vspace{-.4cm}
  \caption{Ratio for system that fires immediately and global
    threshold divided by system with firing delay and prefix-length dependent
    threshold}
  \label{fig:results_fire_delay_prefix_ratio}
  \vspace{-.2cm}
\end{figure}

In general, this confirms RQ5. Our results indicate that prefix-length
dependent thresholds based on
intervals may improve a prescriptive alarm system.

We further consider the use of several prefix-length-dependent thresholds.
\autoref{fig:results_prefix_thresholds_corel} depicts the ranking correlation
between the number of thresholds a system has in ascending order and the
corresponding average cost per case in descending order. As in the previous
experiment, an increasing number of thresholds should not lead to a higher,
average cost per case. If there is no benefit to
additional thresholds, all thresholds could just be set to the same value. This
is not the case in our experiments, though, which we attribute to
overfitting. An example for overfitting are situations, where the basic model
does not fire an
alarm for any case, while the prefix-length dependent approach uses the second
or third threshold to fire alarms for lengthy traces. This leads to better
results on the training set, but not on the test set.

The above results do not generally suggest a positive answer to RQ6. However,
we see consistent improvements or at least no declines for the
non-monotonic versions of the datasets \textit{bpic2017 cancelled} and
\textit{traffic fines}. These datasets have both a balanced class ratio, whereas
the dataset \textit{bpic2017 refused} has an imbalanced class ratio (see
\autoref{table:dataset_stats}). This leads to the conclusion that it is
possible to improve the average cost per case with an increasing number of
thresholds for datasets with a balanced class ratio in non-monotonic cost
scenarios. A limitation to this is that we only observe improvements if the
cost of an undesired outcome is larger than the cost of
compensation. As expected, the improvements in the traffic fines dataset are
higher. Since this dataset has shorter traces, multiple thresholds imply a high
coverage of possible prefix-lengths.

As mentioned in RQ7, we also investigate, if the combination of
prefix-length dependent thresholds and a firing delay is better than the basic
model. To this end, we first check, if the combined approach outperforms a
system with a global threshold and without firing delay. The
results in
\autoref{fig:results_fire_delay_prefix_ratio} confirm that the
proposed approach leads to equivalent or better ratios of average processing
cost under nearly all tested cost configurations. However, for three cost
configurations in the \emph{traffic fines} dataset, we observe negative results.
Exploring these experiments in more detail, we found the negative results
for two of them being due to overfitting. In these scenarios,
the basic model approach builds a system that never fires an alarm, while the
proposed approach builds a system that fires an alarm for some traces. In
the third case, the sophisticated approach sets the fire delay to 1 and sets
the later prefix-length
dependent threshold to 1 and the first threshold to a similar value like the threshold of the basic approach.
With more training runs, therefore, the sophisticated approach would probably
end up with the same behavior as the basic model.

\begin{figure}[t]
  \vspace{-.2cm}
  \centering
  \includegraphics[width=.8\textwidth]{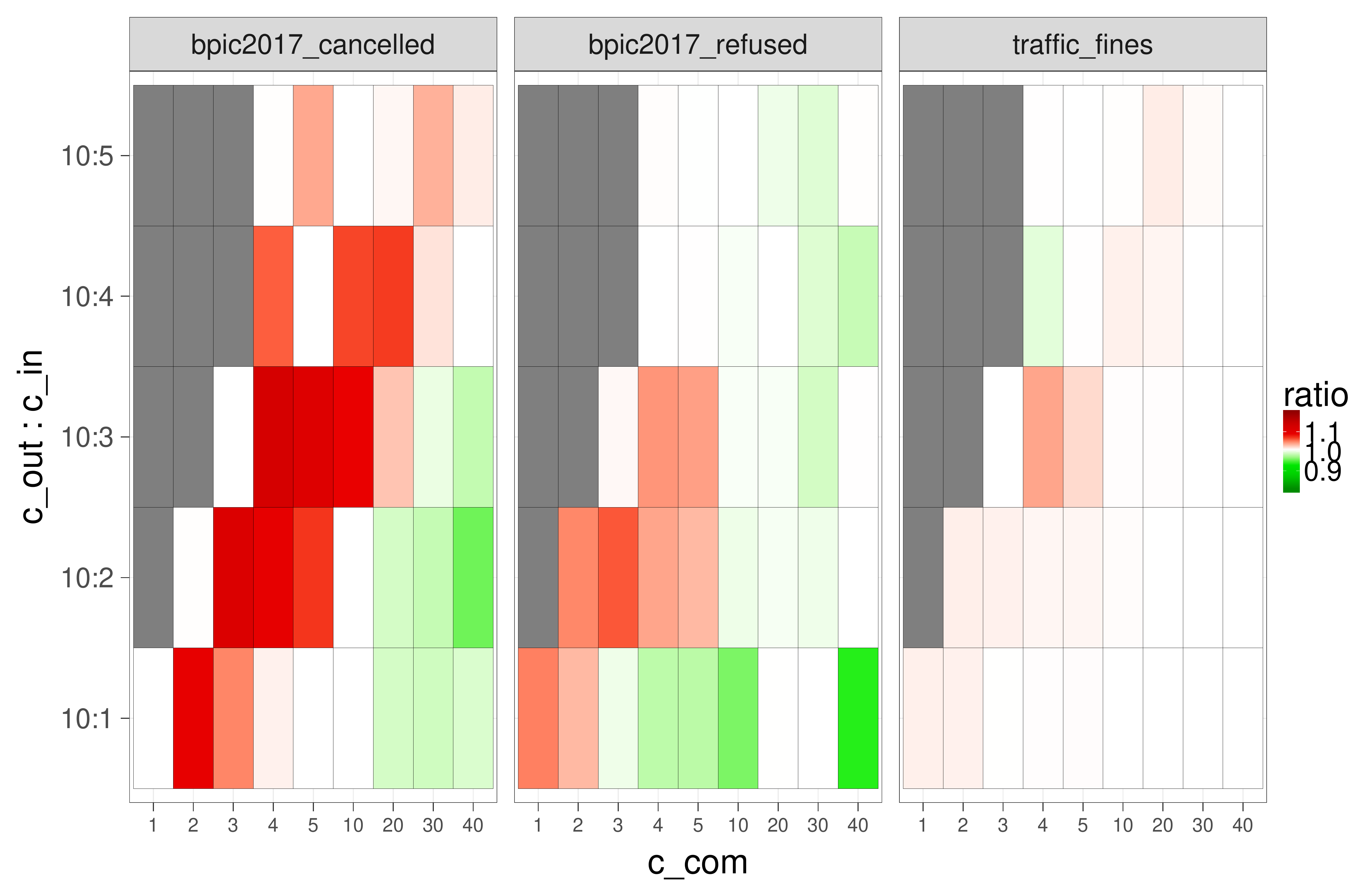}
  \vspace{-.3cm}
  \caption{Ratio of average processing cost for a system with multiple
    alarm based on hierarchical thresholding and the best single alarm system}
  \label{fig:result_1_vs_1_benefit_to_single_hierachical}
  \vspace{-.2cm}
\end{figure}

\begin{figure}[t]
  \vspace{-.2cm}
  \centering
  \includegraphics[width=.6\textwidth]{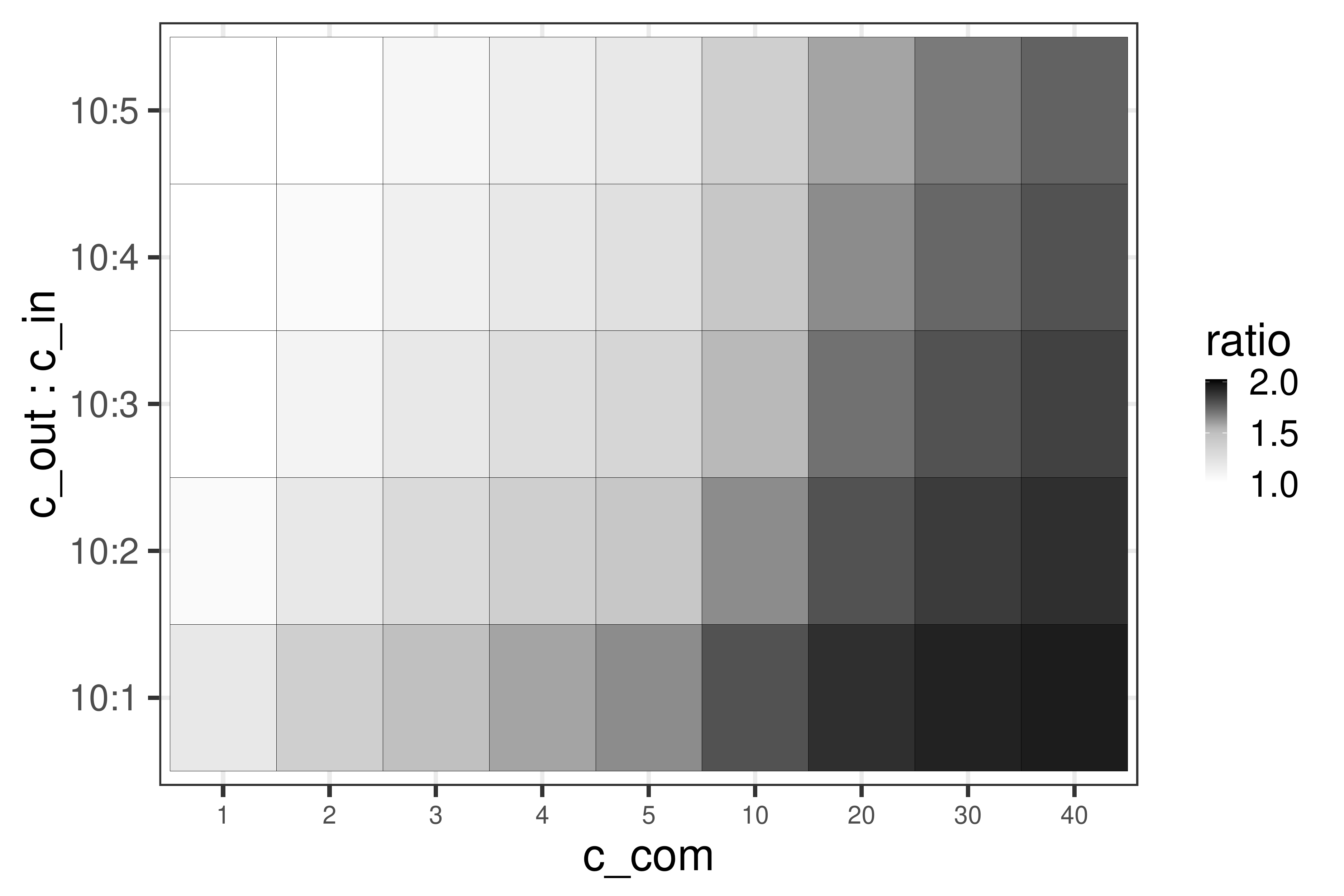}
  \vspace{-.3cm}
  \caption{Ratio of false positives of alarm 1 and alarm 2}
  \label{fig:false_positives_ratio}
  \vspace{-.2cm}
\end{figure}

Finally, we turn to the evaluation of a multi-alarm system (RQ8). Our results
in
\autoref{fig:result_1_vs_1_benefit_to_single_hierachical} indicate that
hierarchical thresholding may, but does not necessarily outperform the
best single-alarm system. Specifically, we observe a stair-like
border for scenarios with high compensation cost, in which hierarchical
thresholding outperforms the best-single alarm system. This aligns with
the relative benefit for false positive for alarm 1 compared to alarm 2. We
visualize this benefit by calculating the ratio of firing alarm 1 for false
positive and alarm 2 in \autoref{fig:false_positives_ratio}. Based thereon, we
conclude that a multi-alarm system using hierarchical
thresholding can outperform the best single-alarm system for scenarios in which
the more expensive alarm, in terms of cost of intervention, has a huge cost
benefit in the case of a false alarm, compared to a cheaper alarm.

A summary of the experimental findings is given in
\autoref{tab:summary_findings}.

\begin{table}[h!]
  \centering  \footnotesize
  \begin{tabular}{l @{\hspace{.5em}} p{8.5cm} p{2cm}}
    \toprule
    RQ & Question & Answer \\ \midrule
    1 & \emph{Does empirical thresholding identify thresholds that consistently
      reduce the average processing cost for different alarm model
      configurations?}&  Yes  \\
    2 & \emph{Does the alarm system consistently yield a benefit over different
      values of the mitigation effectiveness?} &   Yes \\
    3 & \emph{Does the alarm system consistently yield a benefit over different
      values of the cost of compensation?} &  Yes  \\
    \multicolumn{2}{l}{\emph{Is there a reduction in the average processing
    cost
        per case:}} &\\
    4 & \emph{When training a parameter for the minimum number of events that
      exceed the threshold?} & Yes\\
    5 & \emph{When using more than one threshold interval?} & Mostly \\
    6 & \emph{When increasing the number of prefix-length-dependent
    thresholds?} &
    Depends on class ratio and costs\\
    7 & \emph{When combining prefix-length dependent thresholds with a firing
      delay, compared to the a single-alarm system?}& Yes\\
    8 & \emph{Is there a reduction in the average processing cost per case when
      using multiple alarms compared to the best system with only one alarm?} &
    {Depends on cost difference of alarms} \\
    \bottomrule
  \end{tabular}
  \caption{Summary of our findings with respect to the research questions}
  \label{tab:summary_findings}
\end{table}

\section{Conclusion}
\label{sec:conclusion}

This article presented a prescriptive process monitoring framework that extends 
existing approaches for predictive process monitoring with a mechanism to raise 
alarms, and hence trigger interventions, to prevent or mitigate the 
effects of undesired outcomes.
The framework incorporates a cost model to capture the trade-offs between the 
cost of intervention, the benefit of mitigating or preventing undesired 
outcomes, and the cost of compensating for unnecessary interventions. 
We also showed how to optimize the 
threshold(s) for generating alarms, with respect to a given configuration of 
the alarm model and event log.

An empirical evaluation on real-life logs showed significant benefits in 
optimizing the alarm threshold, relative to the baseline case where an alarm is 
raised when the likelihood of a negative outcome exceeds a pre-determined 
value. 
We also highlighted that under some conditions, it is preferable to use 
multiple alarm thresholds. 
Finally, it became apparent that additional cost reductions can be obtained, in 
some configurations, by delaying the firing of an alarm. These findings provide 
insights into how alarm-based systems for prescriptive process monitoring shall 
be configured in practice.

Our work opens various directions for future research. We plan to lift 
the assumption of an alarm always triggering an intervention (regardless of, 
for example, the workload of process workers). To this end, 
our framework shall be extended with notions of resource capacity and 
utilization, possibly drawing on our previous work on risk-aware resource 
allocation across concurrent cases~\cite{conforti2015recommendation}. We also 
strive for incremental tuning of the alarming mechanism based on feedback 
about the alarm relevance and the intervention effectiveness. 
We foresee that active learning methods could be applied in this context.





\section*{Acknowledgement}
This work was supported by the Estonian Research Council (grant IUT20-55) and
the European Research Council (PIX Project).

\bibliographystyle{spmpsci}
\bibliography{bibliography,maxBib}

\end{document}